\pdfoutput=1
\documentclass[journal]{IEEEtran}
\usepackage{hyperref}
\usepackage{framed,multirow}
\usepackage{cite}
\usepackage{amsmath, amssymb}
\usepackage{amsthm}
\usepackage{graphicx}
\usepackage{amsfonts}
\usepackage{longtable}
\usepackage{epsf, epsfig}
\usepackage{epstopdf}
\usepackage{subfigure}
\usepackage{algorithm}
\usepackage{algorithmic}
\usepackage{color}
\usepackage[normalem]{ulem}
\usepackage{booktabs}
\usepackage{adjustbox}
\usepackage{xspace}
\usepackage{doi}
\useunder{\uline}{\ul}{}
\usepackage{cancel}
\usepackage{xargs} 
\usepackage[pdftex,dvipsnames]{xcolor} 
\usepackage{cancel}



\graphicspath{{images/}}

\newcommand{\eat}[1]{}

\newcounter{gaocomm} 
  
\definecolor{blue-violet}{rgb}{0.54, 0.17, 0.89}
\definecolor{mygreen}{rgb}{0.0, 0.5, 0.0}
\definecolor{awesome}{rgb}{1.0, 0.13, 0.32}
\definecolor{bostonuniversityred}{rgb}{0.8, 0.0, 0.0}
\newcommand{\red}[1]{{#1}} 
\newcommand{\blue}[1]{} 

\begin{document}

\title{Graph Decoupling Attention Markov Networks for Semi-supervised Graph Node Classification}

\author{
\thanks{Manuscript received April 28, 2021; revised November 9, 2021 and xx x, 2021; accepted xx xx, 2021. 
(Corresponding author: Jian Pu. \textsuperscript{$\star$} indicates equal contribution.)}
\thanks{J. Chen, S. Chen, and J. Zhang are with Shanghai Key Lab of Intelligent Information Processing and the School of Computer Science, Fudan University, Shanghai 200433, China (email: \{chenj19, chensz19, jpzhang\}@fudan.edu.cn).}
\thanks{J. Pu is with the Institute of Science and Technology for Brain-Inspired
Intelligence, Fudan University,
Shanghai, 200433, China (e-mail:
jianpu@fudan.edu.cn)}
\thanks{M. Bai and J. Gao are with the University of Sydney Business School, the University of Sydney, Darlington,
NSW, 2006, Australia (e-mail:
\{mingyuan.bai, junbin.gao\}@sydney.edu.au)}

\normalsize{
Jie Chen\textsuperscript{$\star$} ,
Shouzhen Chen\textsuperscript{$\star$},
Mingyuan Bai,
Jian Pu\textsuperscript{$\ast$},
Junping Zhang,~\IEEEmembership{Member, IEEE}, 
Junbin Gao
}
}

\markboth{IEEE TRANSACTIONS ON NEURAL NETWORKS AND LEARNING SYSTEMS, VOL. XX, NO. XX, XXX 2021}%
{Shell \MakeLowercase{\textit{et al.}}: Bare Demo of IEEEtran.cls for Journals}

\maketitle
\begin{abstract}
    Graph neural networks (GNN) have been ubiquitous in 
    graph node classification tasks.
    Most of GNN methods update the node embedding iteratively by aggregating its neighbors' information. However, they often suffer from negative disturbance, due to edges connecting nodes with different labels.
    One approach to alleviate this negative disturbance is to use attention \red{to learn the weights of aggregation}, but current attention-based GNNs only consider feature similarity and also suffer from the lack of supervision. In this paper, we consider the label dependency of graph nodes and propose a decoupling attention mechanism to learn both hard and soft attention. The hard attention is learned on labels for a refined graph structure with fewer inter-class edges, so that the aggregation's negative disturbance can be reduced.
    \red{The soft attention aims to learn the aggregation weights based on features over the refined graph structure to enhance information gains during message passing.}
    \red{Particularly, we formulate our model under the EM framework, and the learned attention is used to guide the label propagation in the M-step and the feature propagation in the E-step, respectively.}
    Extensive experiments are performed on \red{six} well-known benchmark graph datasets to verify the effectiveness of the proposed method.
\end{abstract}


\section{Introduction}\label{Sec:1}
Node representation learning on graphs aims to extract high-level features from the node and its neighborhood. It has been proved useful for research areas, such as social influence~\cite{qiu2018deepinf,tnnls-social}, knowledge graphs~\cite{tnnls-kg,park2019estimating}, chemistry and biology~\cite{do2019graph}, and recommendation systems~\cite{ying2018graph,tnnls-rec}. 
Recently, various Graph Neural Networks (GNNs)~\cite{GCN,GAT,tnnls-compre} emerge to solve the node classification problem and achieve state-of-the-art results. Most of them are formulated under the message passing framework~\cite{gilmer2017neural}. 
Each node passes and aggregates messages to/from its neighbors \red{according to edges} to achieve information gain and update its embedding~\cite{ dai2018learning,tiezzi2021deep,hou2020measuring}.

However, the information gain of message passing is not always beneficial in the node classification task, because edges of real-world graphs often connect nodes with different labels. According to~\cite{li2018deeper}, Graph Convolution Networks (GCNs) are regarded as Laplacian smoothers on features. Repeatedly applying the graph convolution operator may smooth the features of the same-label vertices if they are densely connected under the graph structure. Hence the difficulty of the subsequent classification task is significantly reduced. 
However, the graph structure can be noisy from many inter-class edges (which connect two nodes with different labels) \cite{hou2020measuring, wang2020unifying}.
In this case, the graph convolution operator may introduce negative disturbance which arises from neighbors with different labels~\cite{hou2020measuring,xie2020gnns, wang2020unifying}. The reason is that it smooths the features with different labels and blurs the classification boundary. 
Thus, for the node classification task, simply aggregating features 
under the original graph structure
often cannot achieve the optimal performance\cite{hou2020measuring}. 
\red{Taking account of different contributions from the nodes in a graph is important, as not all edges have equal impacts. A wiser solution is to learn the edge weights for message passing in the training stage. }
Furthermore, the labels' information that influences the aggregation quality, \red{should also be considered in the learning process}.

One approach to alleviate the negative disturbance is to leverage the attention mechanism~\cite{vaswani2017attention,hou2020measuring}. It learns the aggregation weights of edges 
\red{based on features}
to choose the critical message and guide the feature propagation~\cite{GAT}.
However, current attention mechanisms on graphs suffer from the lack of supervision~\cite{wang2019improving}.
For example, for citation networks, the weights learned by attention are highly dataset-dependent and degenerated to near-uniform~\cite{SCAGNN}.
Besides, the attention learned using those methods is usually proportional to feature similarity and omits label dependency.
To solve these issues, in this paper, we argue that the attention learning on graphs should be decoupled into the structure learning on labels and the edges' weights learning on features, where more constraints should be applied to learn meaningful attention patterns.

\begin{figure}[t]


\subfigure[Structure influence]{
\begin{minipage}[t]{0.435\linewidth}
\centering
\includegraphics[width=\linewidth]{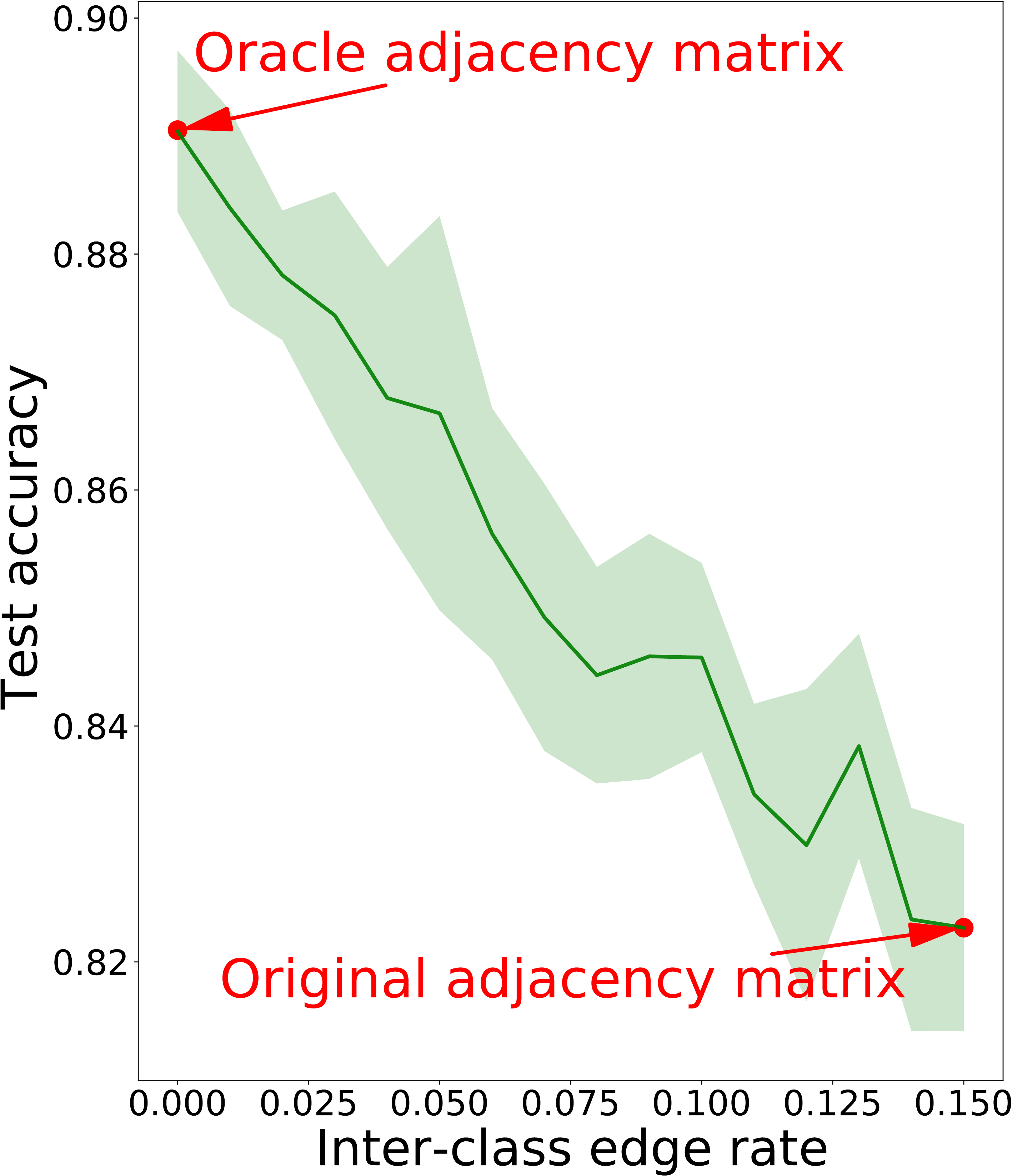}
\label{inter-class curve}
\end{minipage}
}
\subfigure[Weights influence]{
\begin{minipage}[t]{0.46\linewidth}
\centering
\includegraphics[width=\linewidth]{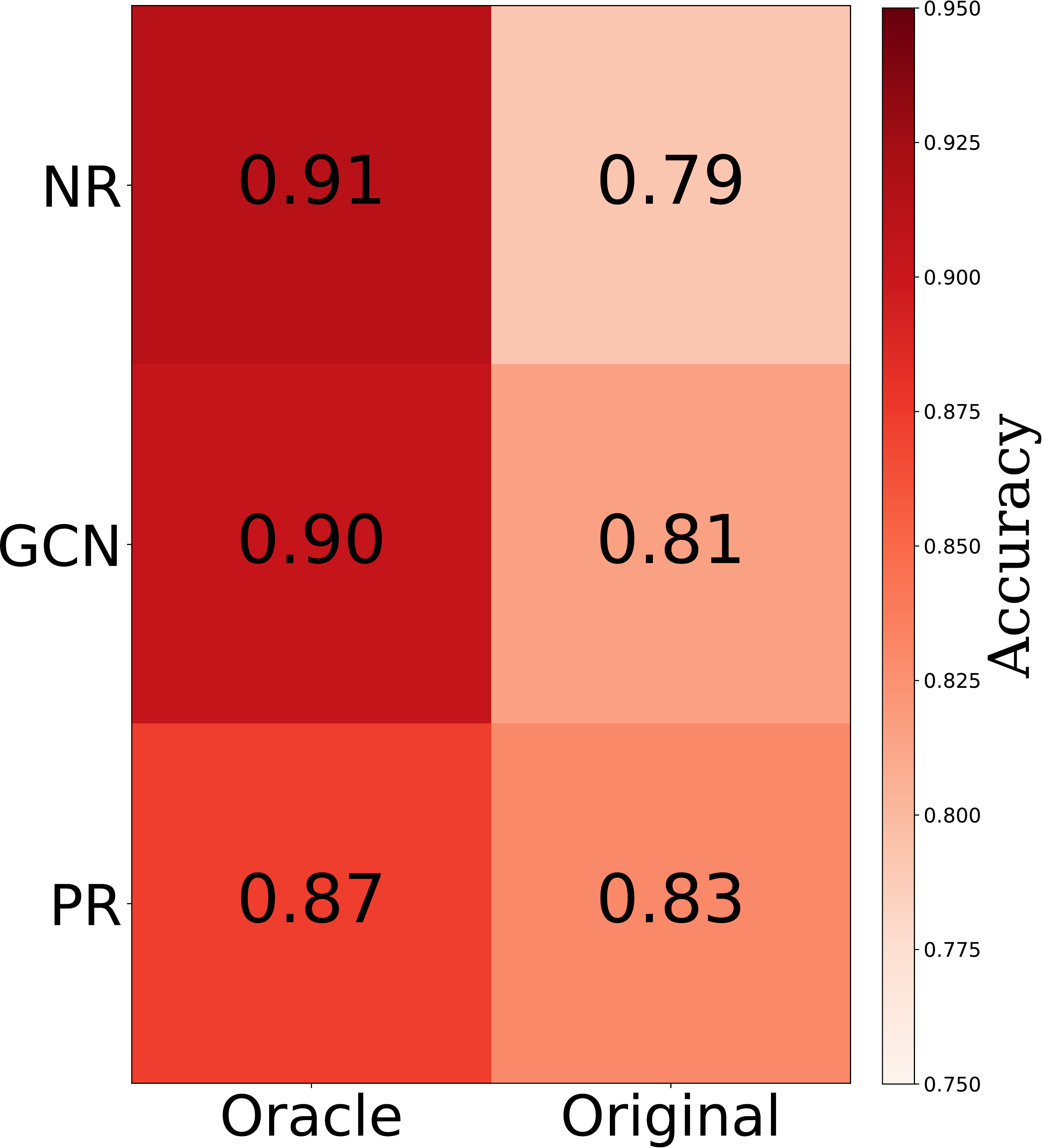}
\label{different_attn_acc}
\end{minipage}
}
\caption{Prediction performance on the Cora dataset using a vanilla GCN trained using (a) adjacency matrices with different inter-class edge rates, a smaller ratio of inter-class edges has smaller negative disturbance and results in higher accuracy for semi-supervised node classification; (b) different attention weights mechanisms. PR is positive relativity attention and NR is negative relativity attention. NR outperforms PR under the oracle adjacency matrix because it can obtain more information gain by paying more attention to dissimilar intra-class neighbors. }
\label{structure_weights}
\end{figure}

Based on the previous studies, by conducting validation experiments in Fig.~\ref{structure_weights}, we demonstrate how the \red{graph} structure and \red{edge} weights affect message passing. We first investigate how the prediction performance varies with the ratio of inter-class edges in the graph structure.
In Fig.~\ref{inter-class curve}, we observe that with the increment of
the ratio of inter-class edges, the test accuracy decreases almost monotonically. It confirms that nodes will receive negative information from their inter-class neighbors (whose labels are different from these nodes) for classification tasks.
Hence, message passing between inter-class nodes should be avoided. 
Next, we examine how performance varies with different aggregation weights. Using vanilla GCN as a baseline, we compare two opposite attention mechanisms, i.e., positive relativity attention (PR) and negative relativity attention (NR). The 
PR
is proportional to $\cos (\mathbf{x}_{i}, \mathbf{x}_{j})$ and the 
NR
is proportional to $ - \cos (\mathbf{x}_{i}, \mathbf{x}_{j})$, where $\mathbf{x}_{i}, \mathbf{x}_{j}$ are the features of nodes $i$ and $j$, respectively. As shown in Fig.~\ref{structure_weights}(b), for the results using the oracle adjacency matrix (all inter-class edges are manually removed), NR is evidently better than PR. While using the original adjacency matrix, the trend is reversed, and PR is obviously better than NR. It indicates that NR is more helpful for message passing under the oracle graph because it pays more attention to divergent intra-class neighbors to obtain more information gain. However, when the inter-class connection exists in the original graph, the negative disturbance is more involved for NR and surpasses its benefits due to the natural difference between different classes.

Thereby, a more flexible attention model for graph node classification consists of two parts, i.e., a refined {\it graph structure} reflecting label similarity (which has fewer inter-class edges)
and reasonable edge {\it weights} (which indicates the aggregation weights for message passing).
Furthermore, the learning mechanisms of structures and weights are different. To obtain a refined graph structure, we need to use nodes' label information to reduce the inter-class edges ratio of the graph. With a refined graph at hand, reasonable weights can be effectively learned by features to avoid negative disturbance and enhance the information gain of aggregation.

In this work, we consider the label dependency and propose the Graph Decoupling Attention Markov Networks (GDAMNs) to decouple the attention learning on the graph for structure learning and weights learning.
To learn {\it a better graph structure}, we model the uncertainty of edges with variational inference and use the hard attention mechanism on node labels to encode the graph structure as a latent variable.
As in the aforementioned validation experiment, we observe that a better graph structure has fewer inter-class edges. Instead of directly using the pseudo-labels to identify inter-class edges, we treat the distribution of node labels as latent variables and evaluate the pseudo-labels iteratively under the expectation-maximization (EM) framework. Furthermore, we impose a novel graph structure prior based on labels and update it in each EM iteration to assist hard attention learning. 
To learn {\it better weights}, we apply the soft attention mechanism on node features over the refined graph structure. We also propose a mutual information constraint as extra supervision on soft attention learning, in order to learn useful weights and enhance information gain. Moreover, the learned attention guides the label propagation in the M-step and the feature propagation in the E-step.

The contributions of this work are summarized as follows:
\begin{enumerate}
\item We decouple the attention learning procedure on the graph to a hard attention graph structure learning on labels and a soft attention graph weight learning on features. We propose to learn a better attention pattern with the constraints which assist to maximize information gain.

\item We define a novel latent graph structure {\it prior} over the label distribution and utilize variational inference to obtain a better graph structure with fewer inter-class connections.

\item We formulate the proposed method under the EM framework. The better graph structure and aggregation weights learned in M-step label propagation guide the feature propagation of E-step to learn the node representation and inference the pseudo-labels distribution. 

\item Extensive experiments are designed and performed to verify the efficiency and superiority of our model and the learned attention.
\end{enumerate}

The remaining part of the paper is organized as follows. Section~\ref{Sec:2} briefly reviews the related methods. Section~\ref{Sec:3} introduces the notations and related graph networks. In Section~\ref{Sec:4}, we present the proposed GDAMN method in detail. Evaluation results on six benchmark datasets and ablation studies are presented in Section~\ref{Sec:5} to verify the effectiveness and robustness of the proposed method. The final section concludes this paper.

\section{Related Work}\label{Sec:2}


There is past literature recognizing the negative disturbance in message passing. Li \textit{et al.}\cite{li2018deeper} proved that the graph convolution operator is actually a special form of Laplacian smoothing. To overcome the negative disturbance, they proposed to combine co-training and self-learning methods to train GCN.  
Moreover, Hou \textit{et al.}\cite{hou2020measuring} discovered that the neighborhood will provide both positive information and negative disturbance for a given task during propagation. 
They proposed smoothness metrics to selectively aggregate neighborhood information to amplify useful information and reduce negative disturbance over the original graph. 
Besides, Xie \textit{et al.}\cite{xie2020gnns} found that neighborhood aggregation may be harmful or unnecessary in some cases. If a node's neighbors have high entropy, the further aggregation will jeopardize model performance, and any more aggregation is unnecessary if a node is nearly identical to its neighbors. 
However, these methods do not study that the graph structure and weights may have different influences on message passing. Unlike them, we decouple the graph structure and influence of weights on message passing and utilize different learning mechanisms to overcome the negative disturbance and enhance the information gain.

To improve the effectiveness of message passing, there are attempts combining learnable attention functions to assign an importance weight to every neighboring node according to the node feature similarity. 
Graph Attention Networks (GATs) \cite{GAT} apply multi-head self-attention mechanisms to learn the aggregation weight distribution on edges among nodes and update node features, and achieve remarkable performance. 
Because of such a successful application of attention mechanisms on GNNs, an in-depth critical study of it becomes a new trend. 
Li \textit{et al.}\cite{SCAGNN} showed that attentions learned by GATs are highly dataset-dependent and the distributions across heads and layers are nearly uniform for all citation networks. 
It is difficult to truly capture the effective weights  that are helpful to the tasks. 
Wang \textit{et al.}\cite{wang2019improving} indicated that GATs suffer from over-fitting due to the increasing number of parameters and the lack of direct supervision on attention weights. They proposed margin-based constraints to reduce information propagation between nodes belonging to different classes. 
However, in most current attention-based approaches, the learned aggregation weights are based on node features, and they fail to capture the useful information provided by label dependency. Hence we instead take both feature and label information to learn better attention patterns and propose an extra mutual information constraint to enhance the supervision.

There are other methods that consider label dependency on graphs. Label Propagation Algorithm (LPA) assumes that two nodes linked by an edge are more likely to have the same label, and encourages the prediction distribution to be equal to a weighted sum of its neighbor distributions \cite{zhu2005semi, tnnls-lpa}.
Besides, Wang \textit{et al.} \cite{wang2020unifying}
\red{\blue{[R3.1]} analyzed the theoretical relationship between LPA and GCN in terms of feature/label smoothness and influence. They proposed a GCN-LPA model by applying the LPA as regularization on training set labels to learn aggregation weights. 
Huang \textit{et al.} \cite{huang2020combining} used a post-process mechanism called C\&S to learn the residual between the ground truth labels and the basic model's prediction. Specifically, they proposed the Autoscale and FDiff-scale strategies to enhance the label propagation to correct and smooth the prediction of basic models.
However, since these methods only consider the label dependency under the original graph structure, they are limited by the number of few known labels under the setting of semi-supervised learning.
There are other works explicitly using GNNs to propagate labels.
Rossi \textit{et al.}~\cite{rossi2018inductive} present an inductive–transductive learning scheme based on GNNs and enrich the node features with the target label in the diffusion process.
}
Graph Markov Neural Networks (GMNN) \cite{qu2019gmnn} used another GCN to update each node label with the labels of its neighbors non-linearly.
The key idea of GMNN is to regard the unlabeled nodes as latent variables and utilize the EM framework to infer better pseudo-labels iteratively. 
\red{However, they neglect the fact that a better graph structure is helpful for the propagation process. To best of our knowledge, the proposed GDAMNs are the first work to simultaneously consider the label influence for structures and the feature influence for weights.}

\section{background}\label{Sec:3}

\subsection{Notations and Problem Setting}

Let $\mathcal{G}=({\mathcal{V}},{\mathcal{E}},\mathbf{X})$ be a graph with the vertex set ${\mathcal{V}} = \{v_1, ..., v_N\}, N = |\mathcal{V}|$. Let $v_i, v_j\in\mathcal{V}$ and the edge set is denoted as ${\mathcal{E}}$ where $e_{ij}=(v_i, v_j)\in\mathcal{E}$. 
Each vertex $v_i$ corresponds to a $d$-dimensional feature representation $\mathbf{x}_{i} \in\mathbb{R}^d$ ($i=1,\cdots, N$), and denote $\mathbf{X}=[\mathbf{x}_1, \cdots, \mathbf{x}_N]$.
Each $\mathbf{y}_{i}$ corresponds to a $c$-dimensional one-hot label representation $\mathbf{y}_{i} \in\mathbb{R}^c$ for vertex $v_i$, and denote $\mathbf{Y}=[\mathbf{y}_1, \cdots, \mathbf{y}_N]$. 
Besides, we use $\hat{\mathbf{Y}}$ to denote the distribution of pseudo-labels.
The edge set ${\mathcal{E}}$ can also be represented by an adjacent matrix ${\mathbf{A}}\in\mathbb{R}^{N\times N}$: 
${\mathbf{A}}_{i,j} = 1$ if $e_{ij}\in \mathcal{E}$, and ${\mathbf{A}}_{i,j} = 0$ otherwise.
Unless particularly specified, the notations used in this paper are illustrated in Table~\ref{table:notations}.

Given the labels ${\mathbf{Y}}_{\mathcal{L}}$ of the nodes ${\mathcal{L}} \subset {\mathcal{V}}$, our goal is to predict the labels $\mathbf{Y}_U$ of the unlabeled nodes ${\mathcal{U}} = {\mathcal{V}} \setminus {\mathcal{L}}$ by exploiting the graph structure ${\mathcal{E}}$ and the features $\mathbf{X}$ corresponding to all the nodes.

\begin{table}[t]
		\caption{Commonly used notations.}
		\label{table:notations}
		\centering
		\setlength{\tabcolsep}{4pt}
		\begin{tabular} {  l l p{7cm} } \toprule
				\textbf{Notations}& \textbf{Descriptions} \\ \midrule
			    $|\cdot|$ & The length of a set. \\ \hline
				$\odot$ & Element-wise product. \\ \hline
				$\|$ & Concatenation operator. \\ \hline
				$\operatorname{Cat}(\cdot)$ & Categorical  operator. \\ \hline

				$\mathcal{G}$ & A graph. \\ \hline
				$\mathcal{V}$& The set of nodes in a graph.\\ \hline
				$\mathcal{U}$& The unlabeled set of nodes in a graph.\\ \hline
				$\mathcal{L}$& The labeled set of nodes in a graph.\\ \hline
				$v_i$ & A node $i\in \mathcal{V}$. \\ \hline
			    ${\mathcal{E}}$ & The set of edges in a graph.\\ \hline
				$\mathcal{N}$, $\mathcal{N}(i)$ & The neighbors set of all nodes, the neighbors of a node $i$. \\ \hline
				$\mathbf{A}$ & The graph adjacency matrix.  \\ \hline
			
			{$\mathbf{A}^{\text{hard}}$}	 & The latent refined graph structure from hard attention. \\ \hline
				$\mathbf{A}^{\text{soft}}$ & The aggregation weights from soft attention. \\ \hline
			{$\mathbf{A}^{\text{stable}}$}	 & The stable aggregation weights for re-weighting. \\ \hline
				$\mathbf{X} \in \mathbf{R}^{N\times d}$ & The feature matrix of a graph. \\ \hline
				$\mathbf{x}_i \in \mathbf{R}^d$ & The feature vector of the node $i$. \\ \hline
				$\mathbf{H} \in \mathbf{R}^{N \times m}$ & The node hidden feature matrix. \\ \hline
				$\mathbf{h}_i \in \mathbf{R}^{m}$ & The hidden feature vector of node $i$. \\ \hline
				$\mathbf{Y} \in \mathbf{R}^{N \times c}$ & The node one-hot label matrix. \\ \hline
				$\mathbf{y}_i \in \mathbf{R}^{c}$ & The label vector of node $i$. \\ \hline		
				$\hat{\mathbf{Y}} \in \mathbf{R}^{N \times c}$ & The node pseudo-label matrix. \\ \hline
				$\hat{\mathbf{y}}_i \in \mathbf{R}^{c}$ & The pseudo-label vector of node $i$. \\ \hline	
				$\ell$ & The layer index \\ \hline
				$\mathbf{W},\phi,\theta$ & Learnable model parameters. \\  
				\bottomrule
		\end{tabular}
\end{table}
	
\subsection{Message Passing in Graph Convolutional Networks} \label{Sec:3.2}

For each node $v_i \in \mathcal{V}$, we denote $\mathcal{N}(i) = \{j: \mathbf{A}_{i,j}\neq 0\}$ as its neighbor set according to the edge set $\mathcal{E}$ and the adjacency matrix $\mathbf{A}$. For the $\ell$-th layer of a GCN, we use $\mathbf{h}^{\ell}_i$ to represent the embedding of node $i$, and $({\mathbf{W}}^{\ell}$,  $\mathbf{b}^{\ell})$ to denote the weights and the bias, and $\sigma(\cdot)$ to be the non-linear activation function. The general GCN message passing rule at the $\ell$-th layer for node $i$ is usually formulated by two steps:
\begin{align}
    \mathbf{m}^{\ell}_i &= \sum\limits_{j\in\mathcal{N}(i) \cup i} {\alpha}_{ij}\mathbf{h}^{\ell-1}_j, & \text{(Neighborhood aggregate)}
    \label{eqn::feat_agg} \\ 
    \mathbf{h}^{\ell}_i &= \sigma( {\mathbf{W}}^{\ell}\mathbf{m}^{\ell}_i + \mathbf{b}^{\ell}), & \text{(Feature transform)} \label{eqn::feat_transform}
\end{align}
The aggregation weight ${\alpha}_{ij}$ in Equation~\eqref{eqn::feat_agg}
can be computed by using the graph Laplacian 
~\cite{Hamilton2017InductiveRL,GCN} or feature-based attention learned in the training process~\cite{GAT,thekumparampil2018attention} as follows: 
\begin{align*}
    {\alpha}_{ij} &= \frac{1}{\sqrt{\left(\left|\mathcal{N}(i)\right|+1\right) \cdot\left(\left|\mathcal{N}(j)\right|+1\right)}}, 
 & \text{(Laplacian based)}  \\ 
    {\alpha}_{ij} &=\frac{\exp \left(\phi_{\omega}\left(\mathbf{h}_{i}, \mathbf{h}_{j}\right)\right)}{\sum_{k \in \mathcal{N}(i)} \exp \left(\phi_{\omega}\left(\mathbf{h}_{i}, \mathbf{h}_{k}\right)\right)},
 & \text{(Attention based)}  
\end{align*}    
where the function $\phi_{\omega}$ in attention based GNNs can be a non-linear transformation to compute the similarity of two node features, i.e., $\phi_{\omega}=\operatorname{LeakyReLU}\left({\mathbf{a}}^{T}\left[\mathbf{W} \mathbf{h}_{i} \| \mathbf{W} \mathbf{h}_{j}\right]\right)$. $\operatorname{LeakyReLU}$\cite{maas2013rectifier} is an activation function,  
$\|$ is concatenation, and $\mathbf{a}$ is a parameterized vector as in GAT~\cite{GAT}. 

After repeating the message passing procedure for multiple layers, the useful information can be propagated to each node of the entire graph. For a GCN with $L$ layers, the prediction of the unlabeled nodes $i\in {\mathcal{U}}$ is made using a \textit{softmax} classifier to the last layer $\mathbf{h}^L_i$:
\begin{align}
    p(\mathbf{y}_{i}\mid \mathbf{X}) = \operatorname{GCN}(\mathbf{A}, \mathbf{X}) = \operatorname{Cat} (\operatorname{softmax}(\mathbf{h}_{i}^L)),
\end{align}
where the $\operatorname{Cat}(\cdot)$ operator denotes the categorical distribution for labels.


\subsection{Graph Markov Neural Networks (GMNN)}
GMNN~\cite{qu2019gmnn} utilizes two GCNs and the pseudo-likelihood variational EM framework to learn label dependency for the semi-supervised node classification task. 
Because the labels of some nodes are unknown, it is difficult to directly maximize the all nodes likelihood
$\log p_{\phi}(\mathbf{Y}|\mathbf{X})$, i.e., $\log p_{\phi}(\mathbf{Y}_{\mathcal{L}}, \mathbf{Y}_{\mathcal{U}}|\mathbf{X})$ which is parameterized by GNN.
As a remedy, we shall compute $\log p_{\phi}(\mathbf{Y}_\mathcal{L}|\mathbf{X})$ which is marginal over the unknown and the evidence lower bound (ELBO) of the log-likelihood function is considered: 
\begin{equation}
\label{eqn::elbo-gmnn}
\begin{aligned}
 \log p_\phi(\mathbf{Y}_{\mathcal{L}}|\mathbf{X}) \geq  \mathbb{E}_{q_\theta(\mathbf{Y}_{\mathcal{U}}|\mathbf{X})}[
 & \log p_\phi(\mathbf{Y}_{\mathcal{L}},\mathbf{Y}_{\mathcal{U}}|\mathbf{X}) \\
 & - \log q_\theta(\mathbf{Y}_{\mathcal{U}}|\mathbf{X})],
\end{aligned}
\end{equation}
where 
$q_\theta({\mathbf{Y}}_{\mathcal{U}}|\mathbf{X})$ 
is a variational approximation of the posterior $p_\phi(\mathbf{Y}_{\mathcal{U}}|\mathbf{Y}_{\mathcal{L}},\mathbf{X})$. 
The lower bound is alternatively optimized between an M-step and an E-step.

\textbf{In the M-step}, with a fixed variational distribution $q_\theta$, the task is to update a GCN parameterized by $\phi$ by maximizing:
\begin{equation}
\label{eqn::obj-p-likelihood}
\begin{aligned}
\mathbb{E}_{q_\theta(\mathbf{Y}_{\mathcal{U}}|\mathbf{X})}[\log p_\phi(\mathbf{Y}_{\mathcal{L}},\mathbf{Y}_{\mathcal{U}}|\mathbf{X})].
\end{aligned}
\end{equation}

Due to the complex data relations, it is hard to define a full joint distribution. Instead, the Markov property is introduced according to the graph local structure. 
Hence, the likelihood function can be approximated by the following pseudo-likelihood:
\begin{equation}
\label{eqn::pseudo-likelihood}
\begin{aligned} 
    \mathbb{E}_{q_{\theta}\left(\mathbf{Y}_{\mathcal{U}} |{\mathbf{X}}\right)}\left[\sum_{i \in {\mathcal{V}}} \log p_{\phi}\left(\mathbf{y}_{i} | \mathbf{Y}_{\mathcal{N}(i)}, \mathbf{X}\right)\right]
\end{aligned}
\end{equation}
where $\mathcal{N}(i)$ denotes the neighbors of node $v_i$ and $\mathbf{Y}_{\mathcal{N}(i)}$ represents labels of node $v_i$'s neighbors.

To evaluate the above expectation, 
we consider the pseudo-label $\hat{\mathbf{Y}}$ and take samples for ${\hat{\mathbf{Y}}_{\mathcal{U}}}$ according to the current $q_{\theta}$, and other ${\hat{\mathbf{Y}}_{\mathcal{L}}}$ remain as the ground truth labels ${{\mathbf{Y}}_{\mathcal{L}}}$ in the training data.
Thus, the network $p_\phi$ is optimized by maximizing
\begin{equation}
\label{eqn::obj-p}
\begin{aligned}
    \sum_{i \in \mathcal{V}} \log p_{\phi}\left(\hat{\mathbf{y}}_{i} | \hat{\mathbf{Y}}_{\mathcal{N}(i)}, \mathbf{X}\right). \ \\
\end{aligned}
\end{equation}
\red{Hence, in the M-step of GMNN, the GCN $p_\phi$ can also be seen as a label propagator to reconstruct labels.}

\textbf{In the E-step}, a GCN is employed to parameterize variational $q_\theta$ for feature propagation according to the predefined structure. The goal is to update the $q_\theta$ to approximate the posterior distribution
$p_\phi(\mathbf{Y}_{\mathcal{U}}|\mathbf{Y}_{\mathcal{L}},\mathbf{X})$. 
Due to the complicated relational structures between node labels, exact inference is intractable. Therefore, we approximate it with another variational distribution $q_\theta(\mathbf{Y}_{\mathcal{U}}| \mathbf{X})$. 
Specifically, based on the mean-field formulation~\cite{qu2019gmnn,mandziuk2002advanced} in GMNN, the optimal $q_\theta(\mathbf{y}_i|\mathbf{X})$ satisfies:
\begin{equation}
\begin{aligned}
q_\theta(\mathbf{y}_i|\mathbf{X}) \approx  p_\phi(\mathbf{y}_i|\mathbf{\hat{Y}}_{\mathcal{N}(i)},\mathbf{X}).
\label{eqn::approx-q}
\end{aligned}
\end{equation}
\red{Moreover, the network $q_\theta$ can be regarded as a feature propagator to produce pseudo-labels.}

In this paper, we use the EM framework proposed by GMNN to solve the problem of insufficient labels. In subsequence, we can leverage the labels' information to learning the structure in the M-step.


\subsection{Gumbel-Softmax Distribution}
It is desired to have the sampling operator differentiable and make the training process of stochastic neural networks end-to-end. To achieve these goals, 
the reparameterization trick \cite{kingma2013auto} is a ubiquitous
technique with continuous variables. However, when the prior distribution is a discrete random distribution, there is no well-defined gradient. The Gumbel-Softmax distribution \cite{jang2016categorical, maddison2016concrete} is such a method to approximate the Categorical distributions by a continuous distribution. Then the gradients can be calculated via the reparameterization trick easily. The Gumbel-Softmax distribution is defined as
\begin{align*}
    \operatorname{Gumbel}({\mathbb{\alpha}}_{1:K}) = \left(\frac{\exp{\left(\left(\log{\alpha _{k}} + G_k \right) / \tau \right)}}{\sum_{k=1}^{K} \exp{\left(\left(\log{\alpha _k} + G_k \right) / \tau\right)}}\right)_{1:K},
\end{align*}
where the subscript $1:K$ is the index of the random variable vector, $\alpha_k$ is the $k$-th element in this vector and proportional to the Categorical distribution. $G_k$ is a noise sampled from the Gumbel distribution. $\tau$ is the temperature parameter which controls Gumbel-Softmax's `sharpness'. A higher $\tau$ will produce a more uniform one.

\section{The Proposed Method}\label{Sec:4}
In this section, we present the proposed Graph Decoupling Attention Markov Networks (GDAMNs) for semi-supervised graph node classification in detail. Our goal is to decouple the attention learning procedure of message passing into two parts: the hard attention on labels for structure learning and the soft attention on features for edge weight learning. 
\red{Under the semi-supervised learning setting, there are no sufficient labels for structure learning. To make use of the unlabeled nodes, we treat the unseen label as a latent variable and leverage the EM framework to approximate the labels' distribution iteratively. After obtaining the label distribution in the E-step, we apply our decoupling attention procedure for structure and weights learning in the M-step.}

\begin{figure}[t]
\begin{minipage}[t]{1\linewidth}
\centering
\includegraphics[width=\linewidth]{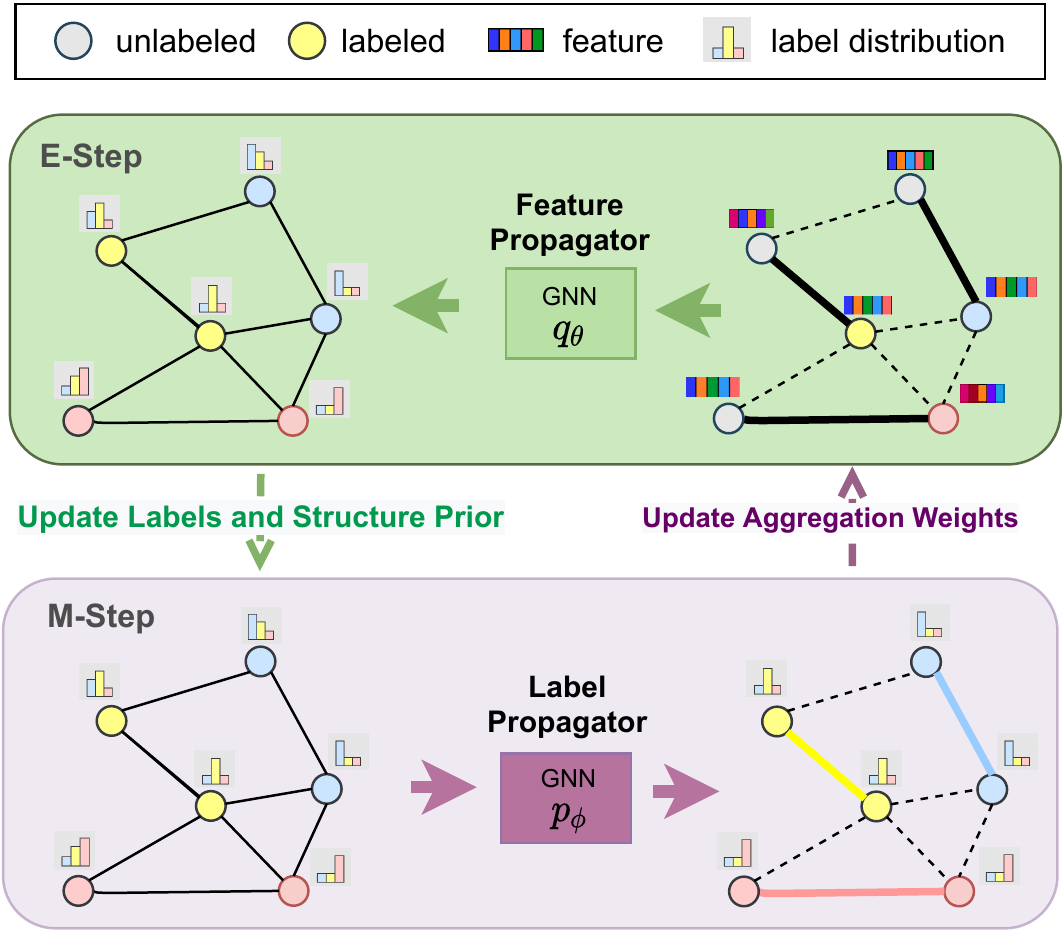}
\end{minipage}
\caption{Overview of the proposed framework: GDAMNs are trained by alternating between the EM steps. In the E-step, the feature propagator $q_{\theta}$ predicts labels by performing feature propagation using the learned aggregation weights by M-step. Then the $q_{\theta}$ passes new labels and the structure prior to the M-step. In the M-step, the label propagation is performed, and the aggregation weights are updated by a decoupling attention on both labels and features. 
Then, $p_{\phi}$ passes the new aggregation weights learned by the attention to the E-step.}
\label{fig:framework}
\end{figure}

As shown in Fig.~\ref{fig:framework}, two GNNs with different structures are parameterized by $\phi$ and $\theta$ respectively, and denoted by $\operatorname{GNN}^{\mathrm{P}}_{\phi}$ and $\operatorname{GNN}^{\mathrm{Q}}_{\theta}$. 
\red{In the E-step, $\operatorname{GNN}^{\mathrm{Q}}_{\theta}$ performs feature propagation according to the aggregation weights learned in the last M-step to predict pseudo-labels. Then, $\operatorname{GNN}^{\mathrm{Q}}_{\theta}$ sends the soft pseudo-labels and the structure prior based on labels to the M-step. In the M-step, $\operatorname{GNN}^{\mathrm{P}}_{\phi}$ performs label propagation to reconstruct the labels.
Moreover, in the $\operatorname{GNN}^{\mathrm{P}}_{\phi}$, the decoupling attention procedure takes both label and feature information to learn the latent graph structure and edge weights. 
Then, the structure and edge weights are transformed into stable aggregation weights for the E-step.}
Note that we perform both feature propagation and label propagation under the learned aggregation weights.
The above procedure helps to reduce the negative disturbance and enhance information gain during the message passing.

Below we first present the variational inference and our structure prior for structure learning in the M-step at Section~\ref{section:graph-inference}. Then we describe our decoupling attention designed for the graph structure and edge weights learning and the mutual information constraint that helps attention learning in the M-step at Section~\ref{section:attention}. Lastly, we show how to re-weight the E-step graph by the aggregation weights learned from the M-step in Section~\ref{section:reweighting}.

\begin{figure*}[t]
    \centering
    \includegraphics[width=1.0\linewidth]{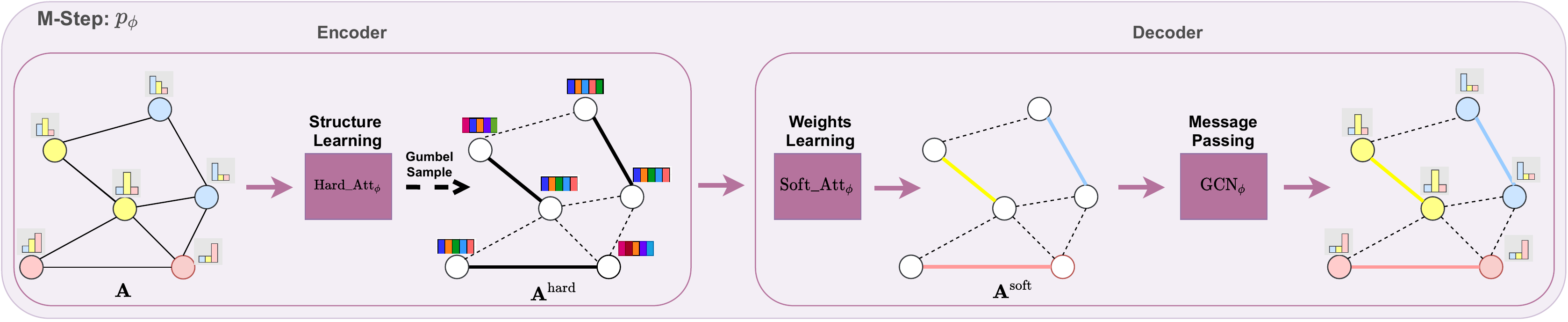}
    \caption{Illustration of the two-phase M-step encoder-decoder architecture: The first phase is the hard attention encoder. The hard attention encodes the 
    refined graph structure
    based on pair-wise labels similarity and is constrained by the KL divergence to the graph prior.
    \red{Then, the $\mathbf{A}^{\text{hard}}$ can be sampled by the Gumbel trick.}
    The second phase is the soft attention and GCN decoder. The soft attention learns the edges' aggregation weights \red{$\mathbf{A}^{\text{soft}}$} based on features under the refined graph structure \red{$\mathbf{A}^{\text{hard}}$} and is regularized by the mutual information constraint. Finally, a simple GCN performs message passing on labels according to the aggregation weights \red{$\mathbf{A}^{\text{soft}}$}.}
    \label{fig:M-step}
\end{figure*}
\subsection{M-Step: Graph Inference and Decoupling Architecture}
\label{section:M-step}
\red{\blue{[R1.1]}In the M-step, we use both the labels and features to learn the hard and soft attention.  
Moreover, for the hard attention and structure learning, instead of directly learning the deterministic structure from labels, we propose a
novel graph prior
to model the uncertainty of edges. Combined with this prior, we use variational inference to take hard attention as the structure encoder to infer better graph structures. For the soft attention and weight learning, we impose neighbor mutual information constraint to learn meaningful weights, and thus reduce effects of negative disturbance.}
\subsubsection{Graph Inference and Label Propagation}
\label{section:graph-inference}
In the M-step, the objective is to update the label propagation parameters $\red{\phi}$ by approximating E-step's pseudo-labels.
On one hand, similar to GMNN~\cite{qu2019gmnn}, we also use the pseudo-likelihood of Markov networks to approximate the likelihood, as in Equation~\eqref{eqn::obj-p}. On the other hand, unlike GMNN~\cite{qu2019gmnn}, we have another goal which is to obtain a new graph according to pseudo-labels' distribution.

As aforementioned in Fig.~\ref{inter-class curve}, the graph structure and similarities of neighbor labels are highly correlated to the 
information gain during message passing. 
Hence, instead of performing the message passing on labels with the original graph structure as GMNN~\cite{qu2019gmnn} does, 
\red{\blue{[R1.1]}
we use the label information to learn a better structure. However, the pseudo-labels may introduce label noises. Directly using the connected nodes' label information to determine the existence of edges is usually suboptimal. So instead of learning a deterministic structure from labels, we use the Bayesian approach to model the uncertainty of structure and introduce a latent variable for a refined graph structure.}
In \cite{kingma2013auto}, the notation of the latent variable is $\mathrm{z}$, whereas in this paper we use $\mathbf{A}^{\text{hard}}$ as the latent variable to denote the refined graph structure. Also, we leverage the variational inference \blue{[R2.1.4]} \red{and a structure encoder $g_\phi$} to infer $\mathbf{A}^{\text{hard}}$ according to the label information 
from the E-step.

In order to infer the new structure  $\mathbf{A}^{\text{hard}}$ based on label information, we consider the ELBO of the log-likelihood function by using the similar derivation developed in conditional variational autoencoder ~\cite{sohn2015learning} and modify \red{the original objective of GMNN's M-step in} Equation~\eqref{eqn::obj-p} as follows:
\begin{equation}\label{eqn::elbo-edge}
\begin{aligned}
    &\quad \log p_{\phi}(\hat{\mathbf{y}}_{i}|\mathbf{\hat{Y}}_{\mathcal{N}(i)}, \mathbf{X}) \\
    &\geq \mathbb{E}_{\mathbf{A}^{\text{hard}} \sim g_{\phi}(\mathbf{A}^{\text{hard}} | \hat{\mathbf{y}}_{i}, \mathbf{\hat{Y}}_{\mathcal{N}(i)}, \mathbf{X})}[\log p_{\phi}(\hat{\mathbf{y}}_{i} | \mathbf{A}^{\text{hard}}, \mathbf{\hat{Y}}_{\mathcal{N}(i)}, \mathbf{X})] \\ &-\operatorname{KL}(g_{\phi}(\mathbf{A}^{\text{hard}} | \hat{\mathbf{y}}_{i}, \mathbf{\hat{Y}}_{\mathcal{N}(i)}, \mathbf{X}), p_{\phi}(\mathbf{A}^{\text{hard}}|\mathbf{\hat{Y}}_{\mathcal{N}(i)}, \mathbf{X})).
\end{aligned}
\end{equation}
\red{Like the normal ELBO, 
the second parameter 
in $\mathrm{KL}(\cdot)$ is our prior distribution $p_{\phi}(\mathbf{A}^{\text{hard}}|\mathbf{\hat{Y}}_{\mathcal{N}(i)}, \mathbf{X})$ of $\mathbf{A}^{\text{hard}}$.
In order to avoid ambiguity and distinguish it from decoder, we omit its footmark $\phi$.} According to the Markov independence property that each node is conditionally independent given its neighbors,
we can summarize the ELBO in Equation~\eqref{eqn::elbo-edge} as below:
\begin{equation}\label{eqn::elbo}
    \begin{aligned}
        O_{\phi,\operatorname{ELBO}} =& \mathbb{E}_{\mathbf{A}^{\text{hard}} \sim g_{\phi}(\mathbf{A}^{\text{hard}} | \mathbf{\hat{Y}}, \mathbf{X})}[\log p_{\phi}(\mathbf{\hat{Y}} | \mathbf{A}^{\text{hard}}, \mathbf{\hat{Y}}, \mathbf{X})] \\ 
        &-\operatorname{KL}(g_{\phi}(\mathbf{A}^{\text{hard}} | \mathbf{\hat{Y}}, \mathbf{X}), \red{p(\mathbf{A}^{\text{hard}}|\mathbf{\hat{Y}}, \mathbf{X})}).
    \end{aligned}
\end{equation}
\red{Note that the above ELBO can also be considered as an encoder-decoder model, i.e.,
the structure encoder 
$g_{\phi}(\mathbf{A}^{\text{hard}} | \mathbf{\hat{Y}}, \mathbf{X})$
encodes the latent graph structure, and the decoder 
$p_{\phi}(\mathbf{\hat{Y}} | \mathbf{A}^{\text{hard}}, \mathbf{\hat{Y}}, \mathbf{X})$
performs message passing under the better graph structure to reconstruct the labels. The implementation details of the encoder 
and the decoder
can be found in Section~\ref{section:attention}. In the following, we describe the expected reconstruction error and the KL divergence of the above ELBO.}

\noindent(a) Reconstruction and Reparameterization:
For the expected reconstruction error in Equation~\eqref{eqn::elbo}, 
the encoder $g_\phi$ use the labels' information to approximate the latent graph structure \red{prior} distribution $p(\mathbf{A}^{\text{hard}}|\mathbf{\hat{Y}}, \mathbf{X})$ with variational parameters predicted by a neural network.
The decoder models the likelihood $p_{\phi}(\mathbf{\hat{Y}} | \mathbf{A}^{\text{hard}}, \mathbf{\hat{Y}}, \mathbf{X})$ to reconstruct labels with a GNN given the latent graph structure $\mathbf{A}^{\text{hard}}$ and the node pseudo-labels distribution over  $\mathbf{\hat{Y}}$.

\blue{[We will remove Equation 11 in the final version, but keep it here to maintain the serial numbers of Equation unchanged.]}
To evaluate the expectation and make the sampling operator differentiable, we utilize the discrete reparametrization trick, i.e., Gumbel-Softmax~\cite{maddison2016concrete,jang2016categorical}, to sample the discrete latent graph structure:
\begin{equation}\label{equ::variational gumbel}
\begin{aligned}
 \mathbf{A}^{\text{hard}} &\sim \operatorname{Gumbel}(g_\phi(\mathbf{A}^{\text{hard}} | \mathbf{\hat{Y}}, \mathbf{X})).
\end{aligned}
\end{equation}



\noindent (b) KL Divergence and the Prior over Graph:
A prior distribution captures our prior belief as to which parameters would have likely generated.
Generally, 
from the earliest probabilistic generative model of graphs developed by
Erd{\H{o}}s \& R{\'e}nyi
\cite{erdHos1960evolution} that assumed an independent identically 
probability for each possible edge, 
we can define the random graph priors on the edges by the independent Bernoulli distribution:
\begin{align}\label{equ::prior}
p(\mathbf{A}^{\text{hard}}) = \prod_{i,j}^{} P(\mathbf{A}^{\text{hard}}_{ij})=\operatorname{Bernoulli} (\rho), \ \ \  
\end{align}
where $\rho$ is the parameter of the Bernoulli matrix distribution, and $P(\mathbf{A}^{\text{hard}}_{ij})$ denotes a $\operatorname{Bernoulli}$ variable representing whether the edge between nodes $v_i$ and $v_j$ exists or not. However, as noted in the previous section, we need to preserve the intra-class edges and eliminate the inter-class edges to obtain a better graph structure for message passing. Thus, instead of making the priors of the edges on graph independent, we can consider the label dependency \red{from label similarity} to make it more flexible and reasonable for message passing. 

\red{\blue{[R1.2]} However, since the labels are discrete, the label similarity for modeling label dependency is less informative, i.e. the similarity of two one-hot representation labels can only be 0 or 1.
As proved in the references of label distribution~\cite{geng2016label} and knowledge distillation~\cite{hinton2015distilling}, the label distribution contains more information than the one-hot label. Thereby, we also use the label distribution (soft label) produced by the softmax of a network output to evaluate the label similarity. Each node is represented by a $c$-dimensional label distribution, and we can compute the cosine similarity between the connected nodes to get the existence prior of each edge. Since the c-dimensional label distribution vector $\mathbf{\hat{y}}_i$ is non-negative and the summation of $\mathbf{\hat{y}}_i$ is 1, the cosine similarity between $\mathbf{\hat{y}}_i$ and $\mathbf{\hat{y}}_j$ ranges from 0 to 1.
Specifically, we define the conditional prior over graphs based on the similarity between label distributions as:
:}
\begin{equation}
    \label{eqn::human-prior}
    \begin{aligned}
    {p(\mathbf{A}^{\text{hard}}_{ij}|\mathbf{\hat{Y}}, \mathbf{X}) = \operatorname{Bernoulli} (\mathbf{A}_{ij} \cos (\mathbf{\hat{y}}_i,\mathbf{\hat{y}}_j)}).
    \end{aligned}
\end{equation}
We set the prior probability of the edges over the original graph positively related to the label similarity between two connected nodes. It encourages the elimination of the harmful edges of the original structure $\mathbf{A}$. Notice that, we update the prior of the latent graph structure in each EM iteration.
Compared with directly using the unreliable pseudo-label to remove inter-class edges, the KL divergence between posterior $g_{\phi}(\mathbf{A}^{\text{hard}} | \mathbf{\hat{Y}}, \mathbf{X})$ and our structure prior $p(\mathbf{A}^{\text{hard}}|\mathbf{\hat{Y}}, \mathbf{X})$ in Equation~\eqref{eqn::elbo} can provide soft supervision and improve the prediction performance.  


\subsubsection{Decoupling Attention Architecture and  Constraint}
\label{section:attention}
\red{In this subsection, we describe the decoupling attention procedure for the implementation of Equation~\eqref{eqn::elbo} in the M-step.}
To maximize the information gain during the message passing process, we should not only consider the label information for structure learning, but also use the feature information for weight learning. 
\red{
So, we first use the hard attention ($\operatorname{Hard\_Att}_{\phi}(\cdot)$) to implement the structure encoder $g_{\phi}$. Then we decompose the decoder into the soft attention weight learner ($\operatorname{Soft\_Att}_{\phi}(\cdot)$) and a simple GCN for information propagation.
Moreover, we impose a mutual information constraint in the attention learning process.}

As shown in Fig.~\ref{fig:M-step}, the $\operatorname{GNN}^{\mathrm{P}}_{\phi}$ in M-step consists of three parts: pair-wise hard attention, soft local attention, and a GCN.
\begin{equation}\label{eqn::m-gnn}
\begin{aligned}
\blue{[R1.3]}
\operatorname{GNN}^{\mathrm{P}}_{\phi} := \underbrace{\operatorname{GCN_\phi} \circ \operatorname{Soft\_Att}_\phi}_{\text{decoder}} \circ \underbrace{\operatorname{Hard\_Att}_\phi}_{\text{encoder}: g_{\phi}},\\
\end{aligned}
\end{equation}
\red{where $\circ$ indicates the composition of operators.} 
\red{The running process of the $\operatorname{GNN}^{\mathrm{P}}_{\phi}$ in the M-step can be divided into the following three steps}:
\begin{equation}\label{eqn::dual attention}
\begin{aligned}
    \mathbf{A}^{\text{hard}} \sim \operatorname{Gumbel}(\operatorname{Hard\_Att}_\phi(\mathbf{A}, \mathbf{\hat{Y}}))&, & \text{(Encoder)} \\
    \mathbf{A}^{\text{soft}} = \operatorname{Soft\_Att}_\phi(\mathbf{A}^{\text{hard}}, \mathbf{X})&, & \text{(Decoder)}\\
·    p_{\phi}(\mathbf{\hat{Y}} \mid \mathbf{A}^{\text{hard}}, \mathbf{\hat{Y}, X}) = \operatorname{Cat}(\operatorname{GCN_\phi}( \mathbf{A}^{\text{soft}}, \left[ \mathbf{\hat{Y}} \| \mathbf{X} \right]))& & \text{(Decoder)} 
\end{aligned}
\end{equation}
where $\mathbf{A}^{\text{hard}}$ and $\mathbf{A}^{\text{soft}}$ denote the refined graph structure with binary and soft aggregation weights, respectively. At first,
we implement the encoder $g_{\phi}$ in Equation~\eqref{eqn::elbo} with $\operatorname{Hard\_Att}_\phi$ to encode the edges' exist probability based on the label distribution $\mathbf{\hat{Y}}$ and the original structure $\mathbf{A}$. Then, we use the Gumbel-Softmax trick to sample a discrete refined structure $\mathbf{A}^{\text{hard}}$ from the edges' exist probality. Once we get the $\mathbf{A}^{\text{hard}}$, we can pass it into the decoder to reconstruct labels. In the decoder, we first apply the $\operatorname{Soft\_Att}_\phi$ to learn the aggregation weights $\mathbf{A}^{\text{soft}}$ based on feature $\mathbf{X}$ and $\mathbf{A^{\text{hard}}}$. Then a simple $\operatorname{GCN_\phi}$ can leverage the \red{aggregation} weights to perform message passing and reconstruct the labels.

In the sequel, we describe the details of $\operatorname{Hard\_Att}_\phi$ and $\operatorname{Soft\_Att}_\phi$. Besides, in order to enhance supervision and the information gain during message passing, we impose a mutual information constraint.

\noindent(a) Hard Attention with Label Similarity:
\blue{R[1.1]}In order to learn a refined graph structure and reduce the negative disturbance, we can use the label information to remove inter-class edges. A natural way is to leverage the ground-truth labels for labeled nodes and pseudo-labels for unlabeled nodes to identify inter-class edges. However, due to limited number of labeled nodes in the semi-supervised setting for real-world datasets, directly using the pseudo-labels may lead to 
\red{incorrect edges' existence}
because of the noisy supervision signals. Aiming to model the uncertainty of edges, we apply variational inference and use the hard attention mechanism on the node label distribution to obtain a better graph structure \red{latent distribution}. 
To further model the similarity of labels, we use the bilinear model~\cite{berg2017graph} to compute the variational probability of edges between node $v_i$ and $v_j$:
\begin{align}\label{eqn::hard-attention}
     \operatorname{Hard\_Att}_\phi(\mathbf{A}, \mathbf{\hat{Y}}) = \mathbf{A} \odot \operatorname{sigmoid}({\hat{{\bf Y}}^T {\bf Q} \hat{{\bf Y}}}),
\end{align}
where $\odot$ denotes the element-wise product, ${\bf Q}\in\mathbb{R}^{C\times C}$ is a \red{\blue{[R2.1.2]}learnable} metric matrix to measure the label similarity of $\hat{{\mathbf{y}}}_i$ and $\hat{{\mathbf{y}}}_j$, and $C$ denotes the number of classes. The hard attention $\operatorname{Hard\_Att}_\phi(\mathbf{A}, \mathbf{\hat{Y}})$ is the implement of the encoder $g_{\phi}(\mathbf{A}^{\text{hard}} | \mathbf{\hat{Y}}, \mathbf{X})$ in Equation~\eqref{eqn::elbo} to infer a refined graph structure based on labels' dependency. 
\red{Notice that we don't need the feature X as input to learning the structure since we assume the label information is enough.}
\red{\blue{[R2.1.3]}As mentioned before, we assume that the existence probability of each edge obeys Bernoulli distribution. So, each element in the output of $g_{\phi}(\mathbf{A}^{\text{hard}}|\mathbf{\hat{Y}}, \mathbf{X})$ is considered as the corresponding parameter of the edge's existence probability distribution.}
Moreover, the hard attention learning is also constrained by the KL divergence of our graph structure prior defined in Equation~\eqref{eqn::human-prior}, which can alleviate the lack of supervision problem of attention and help the hard attention learn a meaningful pattern. 

\noindent(b) Soft Attention with Feature Dissimilarity:
With hard attention $\operatorname{Hard\_Att}_\phi(\mathbf{A}, \mathbf{\hat{Y}})$ and the refined binary graph structure ${\mathbf{A}}^{\text{hard}}$ at hand, we design the soft attention according to the node features \red{for edge weights learning}. As noted in the previous section, our soft attention for message passing is encouraged to 
focus more on
dissimilar neighbors to increase information gain, and the $\operatorname{Soft\_Att}_\phi$ can be implemented as the following: 
\begin{equation}\label{eqn::soft-attention}
\begin{aligned}
    \mathbf{h}_i &= \mathbf{s} \odot \operatorname{ReLU}(\mathbf{W}^{\mathrm{proj}} \cdot \mathbf{x}_i), \\
    \delta_{ij} &= - \cos{(\mathbf{h}_i, \mathbf{h}_j)}, \\ 
    \mathbf{A}^{\text{soft}}_{ij} &=\frac{\mathbf{A}^{\text{hard}}_{ij} \exp(\delta_{ij})}{\sum_{j=1}^{N} \mathbf{A}^{\text{hard}}_{ij} \exp(\delta_{ij})},
\end{aligned}
\end{equation}
\red{We use each element in $\mathbf{A}^{\text{soft}}$ as aggregation weight to guide the message passing in $\operatorname{GCN}_{\phi}$.}
${\mathbf{W}^{\mathrm{proj}}}\in\mathbb{R}^{d\times m}$ is a projection head to reduce the dimension of node feature embedding $\mathbf{x}_i$ and $\mathbf{s}\in\mathbb{R}^{m}$ is a re-scaling parameter for tuning the importance factor of each dimension. We use the negative cosine similarity between the node's hidden state to obtain the dissimilarity score of nodes, and use $\mathrm{softmax}$ on the neighbors of each node to normalize the score, in order to obtain the attention weights on edges.
After we obtain $\mathbf{A}^{\text{soft}}$, 
$\mathbf{A}^{\text{soft}}_{ij}$
can be used to replace the aggregation weights ${\alpha}_{ij}$ in the message passing Equation~\eqref{eqn::feat_agg} for the simple $\operatorname{GCN_\phi}$ in the M-step.
Nevertheless, other formulas to compute attention based on dissimilarity can also be used in the proposed model to implement edge weights.


\noindent(c) Neighbor Mutual Information Maximization:
Besides learning a refined graph to reduce intra-class edges and improve the positive information gain during message passing, we notice that the attention weights degenerate to the average for citation networks due to the lack of supervision problem~\cite{SCAGNN, wang2019improving}. 
As mentioned in the Fig.~\ref{structure_weights}, 
the desired aggregation weights are supposed to focus more on the dissimilar intra-class neighbors. Thus, they are able to enhance nodes' information gain during message passing.
To tackle this problem, apart from the KL supervision signal for structure learning in the M-step, we impose mutual information regularization to prevent the aggregation weights from being uniform.
Because the mutual information of node features is intractable, we employ the labels inferred by $p_\phi$ as an alternative. Denoting the neighbor label distribution by \blue{[R1.4]} \red{$\mathbf{\hat{Y}_\mathcal{N}}$}, we have:
\begin{equation}
    \begin{aligned}
        P(\mathbf{\hat{Y}}) &= \operatorname{GNN}^{\mathrm{Q}}_{\theta}, \ \ \ \ \ P(\mathbf{\hat{Y}}|\mathbf{\hat{Y}_\mathcal{N}}) = \operatorname{GNN}^{\mathrm{P}}_\phi,\\
     I(\mathbf{\hat{Y}},\mathbf{\hat{Y}_\mathcal{N}}) &= H(\mathbf{\hat{Y}}) - H(\mathbf{\hat{Y}}|\mathbf{\hat{Y}_\mathcal{N}}), \\
    O_{\phi,\ \operatorname{ENT}} &= H(\mathbf{\hat{Y}}|\mathbf{\hat{Y}_\mathcal{N}}) = \mathbb{E}_{\mathbf{\hat{Y}}_\mathcal{N}}\left[H(\operatorname{GNN}^{\mathrm{P}}_\phi)\right].
    \end{aligned}
\end{equation}
Here, we use superscript $P$ and $Q$ to indicate the M-step and E-step network, respectively. In the M-step, $P(\mathbf{\hat{Y}})$ and $P(\mathbf{\hat{Y}_\mathcal{N}})$ are fixed, because they are approximated by the fixed network $\operatorname{GNN}^{\mathrm{Q}}_\phi$ in the E-step. 
\red{$P(\mathbf{\hat{Y}}|\mathbf{\hat{Y}_\mathcal{N}})$ indicates the label distribution after the label propagation by $\operatorname{GNN}^{\mathrm{P}}_\phi$. As mentioned before, proper aggregation weights learned by $\operatorname{GNN}^{\mathrm{P}}_\phi$ can help to enhance the information gain and reduce the uncertainty of label distribution.} 
The mutual information $I(\mathbf{\hat{Y}},\mathbf{\hat{Y}_\mathcal{N}})$ computes the reduction in labels \red{uncertainty} after the message passing in the M-step. 
\red{Hence, maximizing the mutual information $I(\mathbf{\hat{Y}},\mathbf{\hat{Y}_\mathcal{N}})$ helps to learn proper aggregation weights for propagating labels.}
When we are optimizing the network $\operatorname{GNN}^{\mathrm{P}}_\phi$ in the M-step using backpropagation, there is no gradient backpropagating to $H(\mathbf{\hat{Y}})$ given $\operatorname{GNN}^{\mathrm{Q}}_\theta$. Therefore, maximizing the mutual information $I(\mathbf{\hat{Y}},\mathbf{\hat{Y}_\mathcal{N}})$ in the M-step is equivalent to minimizing the entropy of $\operatorname{GNN}^{\mathrm{P}}_\phi$.
This entropy term can also be regarded as assisting to learn meaningful aggregation weights that may improve the prediction confidence by integrating beneficial information from neighbors. 
The final objective for M-step is:

\begin{equation}
    \label{eqn::obj-p-final}
    \begin{aligned}
    O_{\phi} = O_{\phi,\ \operatorname{ELBO}}- \beta O_{\phi, \ \text{ENT}}
    \end{aligned}
\end{equation}
where the scalar $\beta\in\mathbb{R}$ is a regularization hyperparameter to weight the mutual information constraint. 


\subsection{E-step: Stable Graph Re-Weighting and Feature Propagation}
\label{section:reweighting}
To calculate the joint distribution expectation in the E-step, as shown in Equation~\eqref{eqn::approx-q}, the task of the E-step is to approximate the posterior distribution of unlabeled nodes $p_{\phi}({\mathbf{Y} }_{\mathcal{U}} | {\mathbf{Y}}_{\mathcal{L}}, {\mathbf{X}})$ by updating the parameters of variational $q_{\theta}({\mathbf{Y}}_{\mathcal{ U}}| {\mathbf{X}})$.
Unlike GMNN~\cite{qu2019gmnn} that $p_{\phi}$ propagates labels under the original structure, we consider the label dependency under a refined graph structure to reduce the negative disturbance. Thus, our posterior prediction of latent labels is the marginal distribution over the latent graph $\mathbf{A}^{\text{hard}}$, and it can be computed by:
\begin{equation}\label{eqn::marnial-infer}
\begin{aligned}
    p({\mathbf{Y}}_{\mathcal{U}} | {\mathbf{Y}}_{\mathcal{L}}, {\mathbf{X}}) &= p_{\phi}({\mathbf{Y}}_{\mathcal{U}} | {\hat{\mathbf{Y}}}, {\mathbf{X}}) \\
     &= \sum_{{\mathbf{A}^{\text{hard}}}} p_{\phi}({\mathbf{Y}}_{\mathcal{U}} | {\mathbf{A}^{\text{hard}}}, {\hat{\mathbf{Y}}}, {\mathbf {X}}) p_{\phi}({\mathbf{A}^{\text{hard}}} | {\hat{\mathbf{Y}}}, {\mathbf{X}}) \\
     & \approx \frac{1}{S} \sum_{s=1}^{S} p_{\phi}({\mathbf{Y}}_{\mathcal{U}} | {\mathbf{A}^{\text{hard}(S)} }, {\hat{\mathbf{Y}}}, {\mathbf{X}})\\
\end{aligned}
\end{equation}
where ${\mathbf{A}^{\text{hard}(S)}}$ is a sample from the posterior graph distribution of the hard attention encoder $g_{\phi}({\mathbf{A}^{\text{hard}}} | {\hat{\mathbf{Y}}}, {\mathbf{X}})$ in the M-step. $S$ is the total number of sampling and is good enough to set as $5$ in practice. 

For the variational $q_{\theta}({\mathbf{Y}}_{\mathcal{ U}}| {\mathbf{X}})$, unlike GMNN~\cite{qu2019gmnn} that propagates features on the original graph, we also perform feature propagation in the E-step according to more reasonable weights learned from the M-step. However, sampling a discrete variable from Gumbel-Softmax may cause a large variance~\cite{jang2016categorical}. If we directly sample the graph structure from hard attention in the M-step and pass it to the E-step, the training process may be unstable and suboptimal. 
Aiming to alleviate the variance, one approach is to sample $S$ times and average the ${\mathbf{A}^{\text{hard}(S)}}$. From the averaged ${\mathbf{A}^{\text{hard}(S)}}$'s, we can get weights for E-step feature propagation. However, it may be expensive when the $S$ is large. To solve these issues, we use a simple yet effective fusion of hard and soft attention to obtain stable weights ${\mathbf A}^{\text{stable}}$ for message passing in the E-step:
\begin{align}\label{eqn::stable-graph}
    {\mathbf A}^{\text{stable}} = \operatorname{Hard\_Att_\phi}(\mathbf{A}, \mathbf{\hat{Y}}) \odot \operatorname{Soft\_Att_\phi}(\mathbf{A}, \mathbf{X}).
\end{align}

The stable weights ${\mathbf A}^{\text{stable}}$ are more reliable for message passing. Note that the soft attention learns weights by node features over the graph, and the hard attention probability can be regarded as a re-scaling factor indicating the label similarity. Hence the fused weights are more stable and
representative in terms of both features and labels. In consequence, unlike  Equation~\eqref{eqn::approx-q} in GMNN, the optimal state of our feature propagator $q_\theta(\mathbf{y}_i|\mathbf{X}, {\mathbf A}^{\text{stable}})$ needs to satisfy:
\begin{align}\label{eqn::optimal-q-edge}
q_\theta(\mathbf{y}_i|\mathbf{X}, {\mathbf A}^{\text{stable}}) \approx  p_\phi({\mathbf{y}}_i | \mathbf{\hat{Y}}, {\mathbf{X}}), \quad i\in {\mathcal{U}}.
\end{align}

For the unlabeled data ${\mathcal{U}}$, we train $q_\theta$ by minimizing the reverse KL divergence between our feature propagator $q_\theta(\mathbf{y}_i|\mathbf{X}, {\mathbf A}^{\text{stable}})$ and the fixed target posterior $p({\mathbf{y}}_i | {\mathbf{\hat{Y}}}, {\mathbf{X}})$ in Equation~\eqref{eqn::marnial-infer}. For the labeled data, we compute the log likelihood directly. Combining these two parts of unlabeled and labeled samples, 
we intend to maximize the following objective function $O_{\theta}$
\begin{equation}\label{eqn::obj-q-final}
\begin{aligned}
    O_{\theta} &=  \sum_{i \in {\mathcal{L}}} \log q_\theta(\mathbf{y}_i|\mathbf{X}, {\mathbf A}^{\text{stable}}) \\ 
&+ \lambda \sum_{i \in {\mathcal{U}}} \mathbb{E}_{p_\phi({\mathbf{y}}_i | \mathbf{\hat{Y}}, {\mathbf{X}})}\left[\log q_\theta(\mathbf{y}_i|\mathbf{X},{\mathbf A}^{\text{stable}})\right],
\end{aligned}
\end{equation}
where $\lambda\in\mathbb{R}$ is a hyperparameter to balance the losses of labeled and unlabeled samples. 

\subsection{Optimization \& Algorithm}
\label{sec::optim}
The detailed algorithm is summarized in Alg.~\ref{alg::optim}. We first train a feature propagator $q_\theta$ using labeled data with the entropy regularizer to encourage the initial pseudo label distribution with the sharper distribution, i.e. the distribution with the large kurtosis. Then we alternatively optimize $p_\phi$ and $q_\theta$ using the EM procedure until convergence. In the M-step, we first update the pseudo-labels by $q_\theta$, and then run the decoupling attention to obtain the better structure $\mathbf{A}^{\text{hard}}$ and weights $\mathbf{A}^{\text{soft}}$ for the message passing. In the E-step, we first update the pseudo-labels by $p_\phi$, and then use the stable reweighting graph ${\mathbf A}^{\text{stable}}$ for the $q_\theta$ to perform message passing on features. The final result is reported using $q_\theta$.

\red{\blue{[R3.2]}In the inference time, we can employe both $p_\phi$ and $q_\theta$ to infer the labels of unlabeled nodes. However, we find that $q_\theta$ consistently outperforms $p_\phi$ in practice.
Thus we use $q_\theta$ with the stable weights ${\mathbf A}^{\text{stable}}$ learned by $p_\phi$ to infer the unlabeled nodes.}
\begin{align}
    \label{q_inference}
    q_{\theta}(\mathbf{Y} \mid {\mathbf A}^{\text{stable}}, \mathbf{X}) = \operatorname{Cat} ( \operatorname{GNN^{\mathrm{Q}}_{\theta}} ({\mathbf A}^{\text{stable}}, \mathbf{X} ) ).
\end{align}

\begin{algorithm}[t]
    \caption{Optimization of the proposed GDAMN method.}
    \label{alg::optim}
    \footnotesize
    \begin{algorithmic}[1]
        \STATE {\bfseries Input:} Graph adjacency matrix ${\mathbf{A}}$, node feature $\mathbf{{X}}$, labeled nodes $\mathbf{Y}_{\mathcal{L}}$, and unlabeled nodes ${\mathbf{Y}}_{\mathcal{U}}$.
        \STATE {\bfseries Output:} Labels $\mathbf{Y}_{\mathcal{U}}$ for the unlabeled sets ${\mathcal{U}}$.
        \STATE Pre-train feature propagator to obtain initial $q_{\theta}$ with ENT regularizer.
        \WHILE{not converge}
            \STATE $\boxdot$ \textbf{M-Step: Graph Inference and Decoupling Architecture}
            \STATE Annotate the unlabeled objects $\hat{\mathbf{Y}}_{\mathcal{U}}$ by $q_\theta$.
            \STATE {Calculate $g_{\phi}(\mathbf{A}^{\text{hard}}|\mathbf{\hat{Y}}, \mathbf{X})$ by employing $\operatorname{Hard\_Att}$ on labels according to Eq.~\eqref{eqn::hard-attention}.}
            \STATE {Sample a discrete graph structure $\mathbf{A}^{\text{hard}} \sim g_{\phi}(\mathbf{A}^{\text{hard}}|\mathbf{\hat{Y}}, \mathbf{X})$ by Gumbel reparameter trick.}
            \STATE {Calculate aggregation weights $\mathbf{A}^{\text{soft}}$ by employing $\operatorname{Soft\_Att}$ based on features and refined structure $\mathbf{A}^{\text{hard}}$ accoding to Eq.~\eqref{eqn::soft-attention}.}
            \STATE {Predict all nodes's label distribution $p_{\phi}(\mathbf{\hat{Y}}|\mathbf{A}^{\text{hard}}, \mathbf{\hat{Y}}, \mathbf{X})$ according to the last equation in Eq.~\eqref{eqn::dual attention}.}
            \STATE {Calculate the M-step optimization object $O_{\phi}$ according to Eq.~\eqref{eqn::obj-p-final} and update parameters $\phi$ by gradient descent.}
            \STATE $\boxdot$ \textbf{E-Step: Stable Graph Re-weighting and Feature Propagation}
            \STATE {According to Eq.~\eqref{eqn::marnial-infer} to inference $p_{\phi}$ for getting the latent graph distribution $\mathbf{A}^{\text{hard}}$ and corresponding $\mathbf{A}^{\text{soft}}$.}
            \STATE {Calculate $\mathbf{A}^{\text{stable}}$ according to Eq.~\eqref{eqn::stable-graph}}
            \STATE {Predict all nodes' label distribution $q_{\theta}(\mathbf{Y}|\mathbf{A}^{\text{stable}}, \mathbf{X})$ according to Eq.~\eqref{q_inference}}
            \STATE  {Calculate the E-step optimization object $O_{\theta}$ according to Eq.~\eqref{eqn::obj-q-final} and update parameters $\theta$ by gradient descent.}
        \ENDWHILE
        \STATE Predict the unlabeled objects $\hat{\mathbf{Y}}_{\mathcal{U}}$ by $q_\theta$.
\end{algorithmic}
\end{algorithm}

\section{Experiments}\label{Sec:5}
In this section, extensive experiments are performed to verify the effectiveness and superiority of our method. To begin with, we report the experimental results on six ubiquitous benchmark graph datasets for node classification. Afterwards, visualized results and ablation studies are discussed to show the effectiveness of learned attention.

\begin{table*}[htp]
\small
	\caption{Statistics of the graph datasets used in the experiments.}
	\label{table:statistics}

	\centering
	\setlength{\tabcolsep}{14pt}
	\begin{tabular}{l|rrrrrr}
		\toprule
		& \textbf{Cora} & \textbf{Citeseer} & \textbf{Pubmed} & \textbf{Coauthor-CS} & \textbf{Coauthor-Phy} & \textbf{OGB-ArXiv}\\
		\midrule
		\# Nodes & 2,708 & 3,327 & 19,717 & 18,333 & 34,493 & 169,343\\
		\# Edges & 5,278 & 4,552 & 44,324 & 81,894 & 247,962 & 1,166,243\\
		\# Features & 1,433 & 3,703 & 500 & 6,805 & 8,415 & 128\\
		\# Classes & 7 & 6 & 3 & 15 & 5 & 40\\
		\# Training Nodes & 140 & 120 & 60 & 300 & 100 & 90,941\\
		\# Validation Nodes & 500 & 500 & 500 & 450 & 150 & 29,799\\
		\# Test Nodes & 1,000 & 1,000 & 1,000 & 17,583 & 34,243 & 48,603\\
		\bottomrule
	\end{tabular}
\end{table*}


\subsection{Datasets}
We first introduce the graph datasets used for node classification.  
Similar to previous studies~\cite{yang2016revisiting,GCN,GAT, shchur2018pitfalls}, we use six benchmark datasets from~\cite{sen2008collective,shchur2018pitfalls,hu2020open} for performance evaluation: three basic citation datasets: Cora, Citeseer, Pubmed, two coauthor datasets: Coauthor-CS and Coauthor-Phy, and one OGB dataset: Arxiv. The detailed statistics of these six datasets are presented in Table~\ref{table:statistics}.

For the basic citation datasets(Cora, Citeseer, Pubmed), nodes correspond to documents; edges correspond to citation links; the sparse bag-of-words are the feature representation of each node. The label of each node represents the document type. 


For the coauthor datasets(Coauthor-CS, Coauthor-Phy), each node represents one author. If authors have coauthored a paper, their corresponding nodes are connected. The paper's keywords are considered as the feature representation of each node. The label of each node indicates the most active field of the author. 

The Arxiv dataset is from the OGB(Open Graph Benchmark)~\cite{hu2020open}, where nodes represent computer science papers on arXiv. Papers are classified into $40$ classes based on the arXiv subject area. Unlike the basic citation datasets that use bag-of-words as nodes' feature, the node features in Arxiv are computed as the average word embedding of all words in the paper.

We use the same data partition and preprocessing as in~\cite{yang2016revisiting, xie2020gnns, hu2020open} for citation datasets and~\cite{shchur2018pitfalls} for coauthor datasets. The evaluation metric is the prediction accuracy of the test nodes. 

\begin{table*}[htb]
\small
	\caption{Comparison of test accuracies on five benchmark datasets. The mean and standard deviation are reported.}
	\label{table:random_split}
	\centering
	\setlength{\tabcolsep}{14pt}
	\scalebox{1}{
	\begin{tabular}{c|ccccc}
		\toprule
		Method & \textbf{Cora} & \textbf{Citeseer} & \textbf{Pubmed} & \textbf{Coauthor-CS} & \textbf{Coauthor-Phy} \\
		\midrule 
		MLP & 58.2 $\pm$ 2.1 & 59.1 $\pm$ 2.3 & 70.0 $\pm$ 2.1 & 88.3 $\pm$ 0.7 & 88.9 $\pm$ 0.7 \\
		LogReg & 57.1 $\pm$ 2.3 & 61.0 $\pm$ 2.2 & 64.1 $\pm$ 3.1 & 86.4 $\pm$ 0.9 & 86.7 $\pm$ 1.5 \\
		LPA \cite{zhu2005semi} & 74.4 $\pm$ 2.6 & 67.8 $\pm$ 2.1 & 70.5 $\pm$ 5.3 & 73.6 $\pm$ 3.9 & 86.6 $\pm$ 2.0  \\
		\midrule
		CS-GNN \cite{hou2020measuring} & 70.1  & 56.0  & 73.6  & -  & - \\
		GCN \cite{GCN} & 81.5 $\pm$ 0.8 & 70.3 $\pm$ 1.5 & 79.0 $\pm$ 0.4 & 91.1 $\pm$ 0.5 & 92.8 $\pm$ 1.0 \\
		GAT \cite{GAT} & 83.0 $\pm$ 0.7 & 72.5 $\pm$ 0.7 & 79.0 $\pm$ 0.3 & 90.5 $\pm$ 0.6 & 92.5 $\pm$ 0.9 \\
		SGC \cite{wu2019simplifying} & 81.8  & 68.9  & 77.5  & 90.8 $\pm$ 0.5  & 93.2 $\pm$ 0.8 \\
		APPNP \cite{klicpera2018predict} & 84.3  & 68.8  & 78.7  & 91.6 $\pm$ 0.7 & 93.3 $\pm$ 0.7 \\
		ALaGCN \cite{xie2020gnns} & 82.9  & 70.9 & 79.6  & 91.0 $\pm$ 0.8 & 92.9 $\pm$ 0.6 \\
		ALaGAT \cite{xie2020gnns} & 85.0  & 71.5 & 78.1  & 90.2 $\pm$ 0.6 & 92.3 $\pm$ 1.0 \\
		GMNN \cite{qu2019gmnn} & 83.4 & 73.1 & \textbf{81.4}  & 91.4 $\pm$ 0.7 & 93.1 $\pm$ 0.6 \\
	\blue{[R1.5,R3.3]}
		\red{GCN-LPA} \cite{wang2020unifying}   & 84.4 $\pm$ 1.3 & 71.8 $\pm$ 2.0 & 79.2 $\pm$ 1.1 & 91.4 $\pm$ 0.7 & 93.5 $\pm$ 0.9 \\
	\blue{[R3.3]}
		\red{GCN-C\&S} \cite{huang2020combining}   & 83.2 $\pm$ 0.6 & 71.8 $\pm$ 0.5 & 79.4 $\pm$ 0.5 & 91.3 $\pm$ 0.8 & 93.1 $\pm$ 0.6 \\
		\midrule
		GMNN(with GAT) & 83.7 $\pm$ 0.8 & 73.3  $\pm$ 0.7 & 80.2 $\pm$ 0.8 & 91.4 $\pm$ 0.6 & 93.2 $\pm$ 0.8 \\
		GDAMN (w/o ENT) & 84.5 $\pm$ 0.7 & 73.5  $\pm$ 0.5 & 80.4 $\pm$ 0.8 & 91.7 $\pm$ 0.5 & 93.6 $\pm$ 0.8 \\
		GDAMN (w/o Hard) & 84.0 $\pm$ 0.6 & 73.4  $\pm$ 0.5 & 80.3 $\pm$ 0.9 & 91.9 $\pm$ 0.5 & 93.8 $\pm$ 0.7 \\
		GDAMN(w/o Soft) & 84.6 $\pm$ 0.8 & 73.9    $\pm$ 0.9 &  80.7 $\pm$ 0.7 & 91.8 $\pm$ 0.6 & 93.9 $\pm$ 0.6 \\
		GDAMN (ours) & \textbf{85.4 $\pm$ 0.5} & \textbf{74.4 $\pm$ 0.6} & 80.9 $\pm$ 0.7  & \textbf{92.2 $\pm$ 0.5} & \textbf{94.3 $\pm$ 0.6}\\
		\bottomrule
	\end{tabular}
	}
\end{table*}

\begin{table}[t]
    \caption{Performance on the ogbn-arXiv task measured in terms of classification accuracy along with standard deviations. }
    \label{table:arxiv}
    \centering
	\setlength{\tabcolsep}{12pt}
    \begin{tabular}{c|cc}
      \toprule
		Method & \textbf{Validation} & \textbf{Test} \\
      \midrule
        MLP & 57.65 $\pm$ 0.12 & 55.50 $\pm$ 0.23 \\
        GCN\cite{GCN}  & 73.00 $\pm$ 0.17 & 71.74 $\pm$ 0.29 \\
        GraphSAGE\cite{Hamilton2017InductiveRL} & 72.77 $\pm$ 0.16 & 71.49 $\pm$ 0.27 \\
        GMNN\cite{qu2019gmnn}  & 73.17 $\pm$ 0.13 & 71.96 $\pm$ 0.19 \\
        \blue{[R2.3]}\red{DeeperGCN}\cite{li2021deepgcns}  & 72.62 $\pm$ 0.14 & 71.92 $\pm$ 0.16 \\
        \blue{[R2.3]}\red{DAGNN}\cite{liu2020towards}  & 72.90 $\pm$ 0.11 & 72.09 $\pm$ 0.25 \\
        \blue{[R2.3]}\red{ALaGAT}\cite{xie2020gnns}  & 73.22 $\pm$ 0.14 & 72.03 $\pm$ 0.22 \\
        \blue{[R1.5,R3.3]}\red{GCN-LPA}\cite{wang2020unifying}  & 73.13 $\pm$ 0.18 & 71.94 $\pm$ 0.23 \\
        \blue{[R3.3]}\red{GCN-C\&S}\cite{huang2020combining}& 73.33 $\pm$ 0.12 & 72.24 $\pm$ 0.18\\
        GDAMN(ours)   & \textbf{73.48} $\pm$ 0.10 & \textbf{72.35} $\pm$ 0.18 \\
      \bottomrule
    \end{tabular}
\end{table}

\subsection{Baselines}
We compare our method with three non-GNN models and GNN based methods. 
The baseline models are logistic regression (LogReg), multilayer perceptron (MLP), and label propagation (LPA)~\cite{zhu2005semi}. The first two methods are attribute-based models that only use the feature representation of each node for prediction. LPA is a graph-based method that only uses the graph structure for prediction. \red{The other GNNs can be divided into three types:}
\begin{itemize}
    \item \textbf{Classical GNNs}
    \begin{itemize}
        \item \textbf{GCN}~\cite{GCN}: GCN is the pioneer to perform linear approximation to spectral graph convolutions.
        \item \textbf{SGC}~\cite{wu2019simplifying}: SGC reduces GCNs' complexity by removing nonlinearities and collapsing weight matrices between consecutive layers. \red{This method can also be regarded as decoupling the feature transformation and propagation.}
        \item \textbf{APPNP} ~\cite{klicpera2018predict}: APPNP combines GNN with personalized PageRank to separate the neural network from the propagation scheme. 
        \item \textbf{GraphSAGE}~\cite{Hamilton2017InductiveRL}: GraphSAGE learns a function that generates embeddings by sampling and aggregating features from a node’s local neighborhood.
        \item \textbf{GAT}~\cite{GAT}: GAT is a graph neural network that applies the attention mechanism on node feature to learn edge weights.
    \end{itemize}
    \item \blue{[R3.3]} \textbf{Label Information GNNs}
     \begin{itemize}
        \item \textbf{CS-GNN}~\cite{hou2020measuring}:
CS-GNN uses the \red{feature and label smoothness} to help the attention mechanism selectively aggregate neighborhood information and reduce negative disturbance.
        \item \textbf{ALaGCN and  ALaGAT}~\cite{xie2020gnns}: ALaGCN and ALaGAT propose a novel metric and integrate them into an adaptive-layer module to make individual decision at each round of neighborhood aggregation. \red{
        \blue{[R3.3]}
        They estimate whether neighborhood aggregation is harmful or unnecessary by calculating the metric based on label information.}
        \item \textbf{GMNN}~\cite{qu2019gmnn}: GMNN is also an EM-based method, it models the joint distribution of nodes and labels with a conditional random field and 
        \red{\blue{[R3.3]}uses two GCNs to propagate features and labels under the original structure.}
        \item \textbf{GCN-LPA}~\cite{wang2020unifying}:\red{\blue{[R3.3]}GCN-LPA can also be seen as learning attention weights based on training set labels.
However, the input of attention remains node features and it uses the LPA as regularization to assist in learning proper edge weights.}
        \item
        \textbf{GCN-C\&S}~\cite{huang2020combining}:\red{\blue{[R3.3]}C\&S combines the label propagation with a basic predictor to learn the residual. In order for a fair comparison, we choose the GCN as the basic model, since our GDAMN also uses the GCN as a predictor. For the scaling strategy of C\&S, we choose the Autoscale that works more reliably than FDiff-scale, which is also mentioned from their official code.}
    \end{itemize}  
        \item \blue{[R2.3]} \textbf{Deep GNNs}
    \begin{itemize}
        \item \textbf{DeeperGCN}~\cite{li2021deepgcns}:\red{
        \blue{[R2.3]}It combines residual/dense connections and dilated convolutions, and adapts them to GCN architectures to successfully train very deep GCNS.}
        \item \textbf{DAGNN}~\cite{liu2020towards}:\red{
        \blue{[R2.3]}DAGNN decouples the feature transformation and propagation to learn node representation from larger receptive fields. Moreover, it proposes an adaptive adjustment to balance the information from local and global neighbors for each node.}
    \end{itemize}
\end{itemize}

For ease of comparison, we use the main reported results from~\cite{qu2019gmnn, shchur2018pitfalls, xie2020gnns}. Moreover, for the missing results of most baselines on coauthor and arXiv datasets, we rerun their official code over 20 runs. 
\red{\blue{[R1.5,R3.3]}Notice that, for the GCN-LPA and GCN-C\&S, the original papers use the full-supervised data splits which use the train/val/test as 60\%/20\%/20\% for citation and coauthor datasets. For a fair comparison, we also rerun their codes under our semi-supervised data splits that only 20 labeled per class is used for training.}

\subsection{Experiment Settings}
We use 2 hidden layers, and the dimension of hidden layers is set as 16 for citation datasets and 64 for co-author, which are the same as in~\cite{qu2019gmnn,shchur2018pitfalls}. 
We initialize weights according to Kaiming initialization~\cite{he2015delving}, and train our model for 2 EM iterations with 200 epoch per iteration using Adam~\cite{kingma2014adam}. We report the mean test accuracy when the validation accuracy is maximized over 20 experiments with different random seeds. During training, we set the initial learning rate as 0.05, weight decay as (0.0005/0.0001) for (citation/co-author) datasets, and after the first EM iteration the weight decay is set to 0.0001 for citation and co-author datasets. 
For the Arxiv dataset, we set the hidden layers to 3, the hidden dimension to 128, learning rate to 0.02, weight decays to 0 and the number of epochs to 1000 per EM iteration.
We also apply the dropout~\cite{srivastava2014dropout} with the dropout rate $d = 0.5$ on hidden layers and $d = 0.2$ for the attention weights. For the ENT regularization, we select the parameter $\beta\in{\{0.2, 0.4, 0.6, 0.8, 1\}}$ using the validation set. For the weighted parameter $\lambda$ of the cross-entropy loss on the unlabeled data part, we set it to 0.8 for all datasets.


\subsection{Results and Analysis}

\subsubsection{Performance Evaluation}

The performance on the node prediction accuracy is summarized in Table ~\ref{table:random_split} and Table~\ref{table:arxiv}. We first note that almost all GNNs achieve much better prediction accuracy than the non-GNNs models (LogReg, MLP, and LPA). Compared with the GNN-based method that combines graph structure and feature representation of nodes, non-GNNs models merely use partial information for node prediction. Besides, our GDMAN performs better than the Laplacian-based GNNs (GCN, SGC, and APPNP). Because these methods propagate features under the original graph structure and may suffer from the negative disturbance due to intra-class edges.
Moreover, our GDAMN outperforms the attention-based GNNs such as GS-GNN, GAT, and AlaGAT, because our decoupling attention can learn more meaningful attention weights from both features and labels. 
\red{Compared with the label propagation methods (GCN-LPA, GMNN, and GCN-C\&S), our model also achieves better performance in terms of higher prediction accuracies in most datasets. }
Though our performance is slightly worse for the Pubmed dataset than that of the GMNN method, our prediction accuracy is still better than other GNNs. The reason might be that there is a fewer negative disturbance and label dependency in the Pubmed dataset since it has only 3 classes. 


\begin{figure}
\centering
\begin{minipage}[t]{1\linewidth}
\centering
\includegraphics[width=\linewidth]{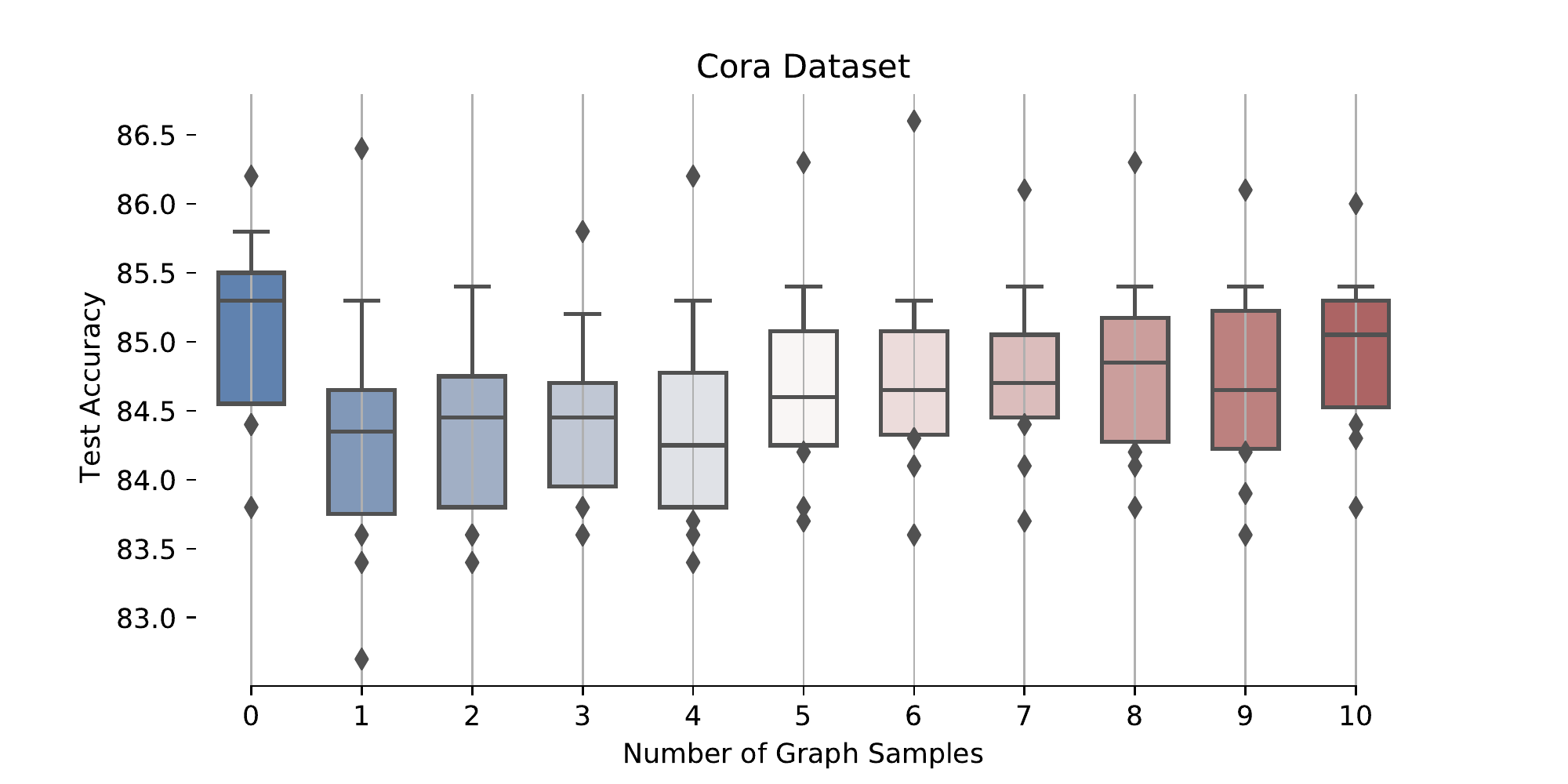}
\begin{minipage}[t]{1\linewidth}
\centering
\includegraphics[width=\linewidth]{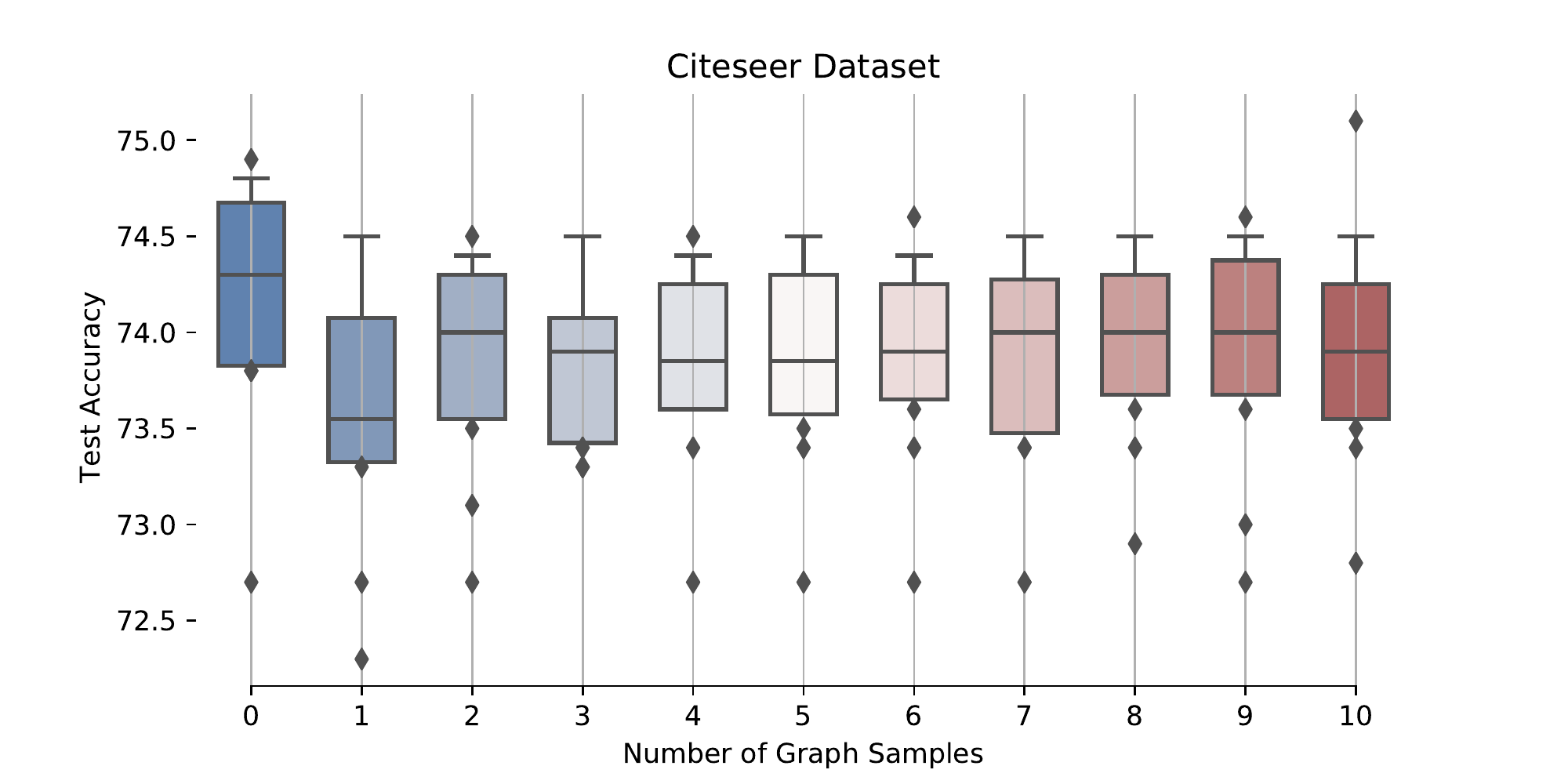}
\end{minipage}
\end{minipage}
\\ 
\caption{From top to bottom, the node classification curve on cora and citeseer datasets under different number of graph samples in the E-step re-weighting. The 0 sample indicates the stable re-weighting and it has higher accuracy than directly sample graph structure from the M-step.}
\label{fig:stable}
\end{figure}

\begin{figure*}[ht]
    \centering
    \includegraphics[width=1\linewidth]{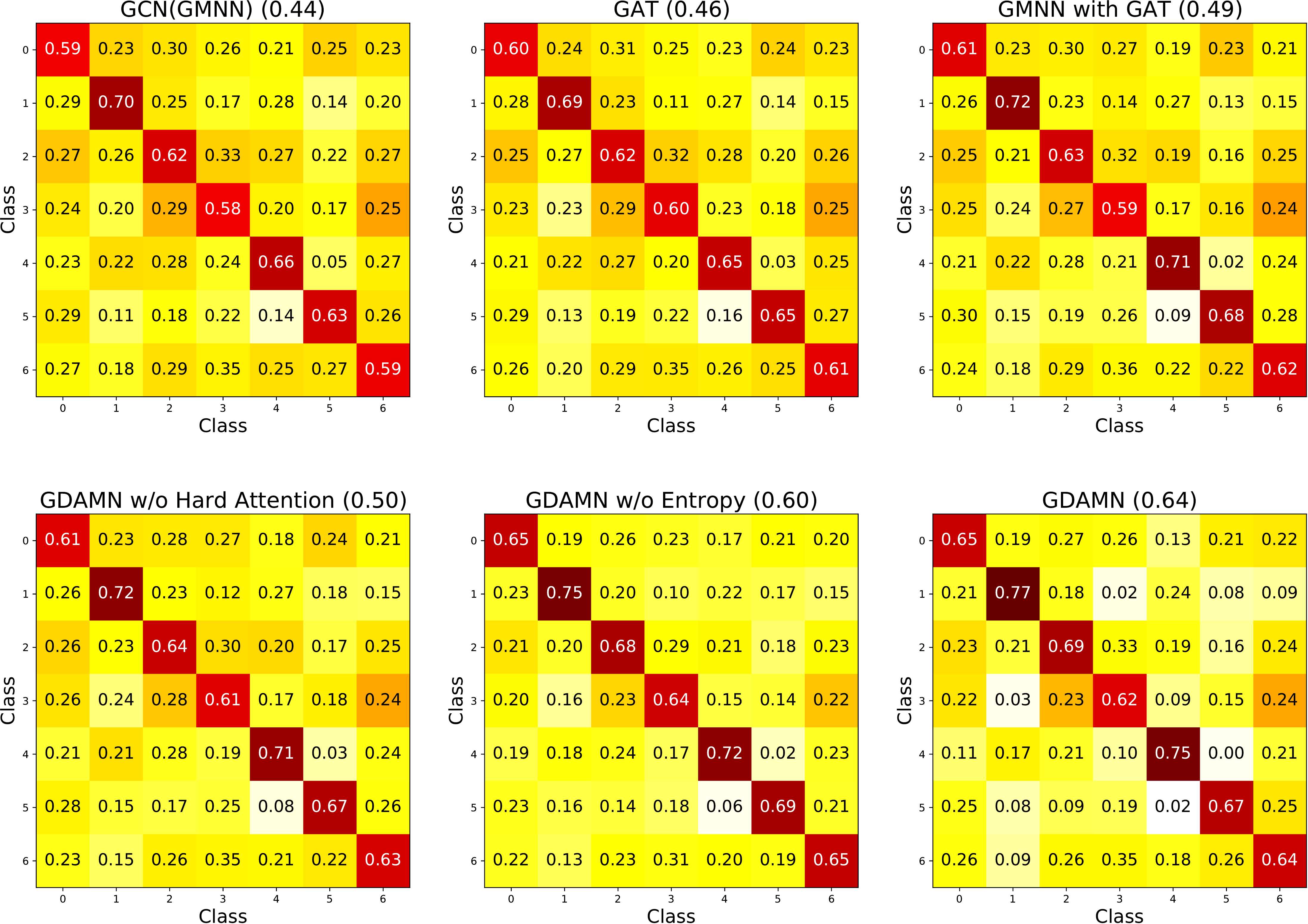}
    \caption{Illustration of the connectivity strength over seven classes of the cora dataset. The scores in brackets are the ratio of DiagSum-to-OffDiagSum of matrix items. A higher score indicates a higher concentration on the diagonal components and a more reliable weights.}
    \label{fig:attention-view}
\end{figure*}

 \begin{figure*}[ht]
     \centering
    \includegraphics[width=0.8\linewidth]{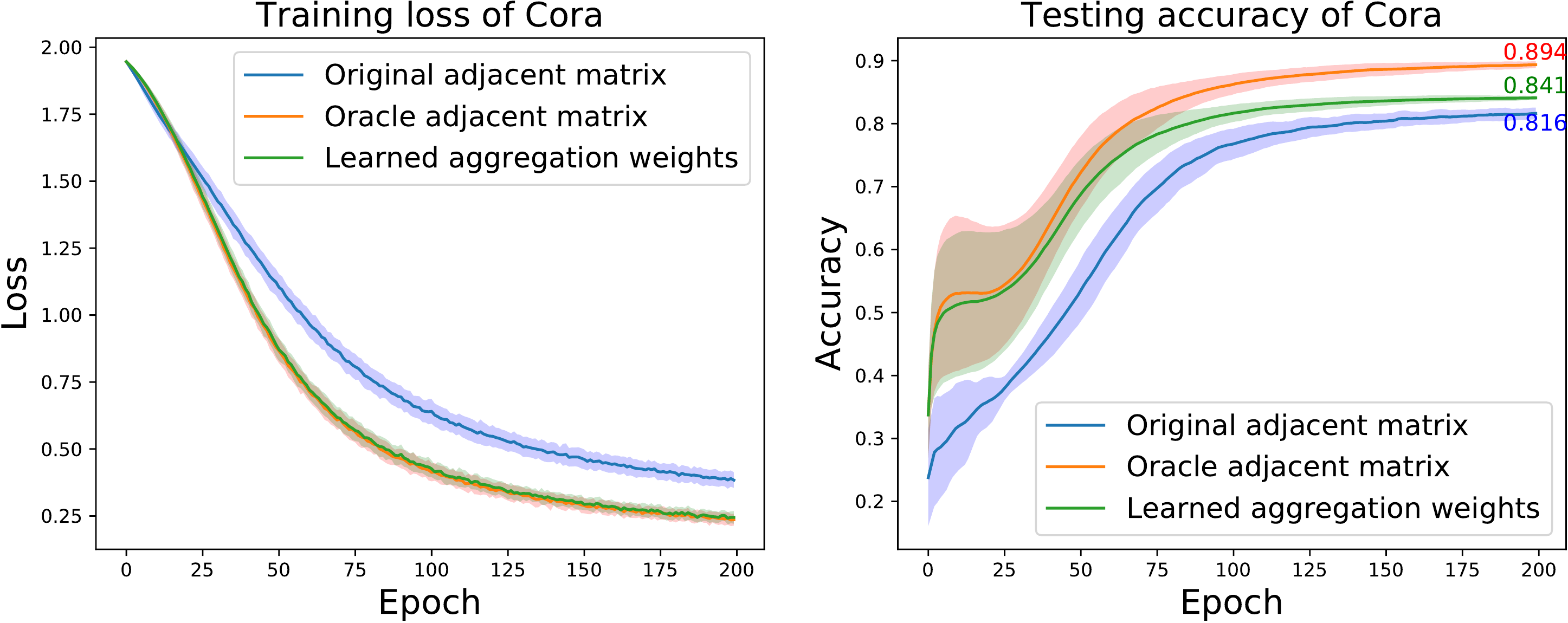}
 	\caption{Performance comparison on the cora dataset using the original adjacent matrix, oracle adjacent matrix, and learned aggregation weights.}
     \label{fig:learn-edge-curve}
 \end{figure*}

\subsubsection{Ablation Study}
In this section, we compare GDAMN with its four variants on all datasets to validate the effectiveness of each component.
\begin{itemize}
\item GMNN (with GAT): a natural variant that we replace the GCN with GAT in GMNN to apply the attention mechanism on labels of nodes. 
\item GDAMN (w/o ENT): GDAMN without the mutual information constraint.
\item GDAMN (w/o Hard): GDAMN without the hard attention and variational inference for structure learning in M-step. 
\item GDAMN (w/o Soft): GDAMN without the soft attention for weights learning in M-step. 
\end{itemize}
As shown in
Table~\ref{table:random_split}, the performance of each variant degrades for all datasets, which demonstrates the effect of ENT regularizer and decoupling two-step attention design. 
The GMNN (with GAT) is slightly better than GMNN. It has shown that perform attention on labels is helpful for message passing. 
The performance degeneration of GDAMN (w/o Hard) and GDAMN (w/o Soft) confirm that both graph structure and weights are important for message passing in node classification tasks. Apart from them, GDAMN (w/o ENT) outperforms GMNN (with GAT) shows that our decoupling attention design is more useful for each node to obtain information gain during message passing.

\subsubsection{The learned Attention}
We then visualize the attention to demonstrate the effectiveness of our learning procedure. As stated in Introduction, we expect the intra-class connectivities to be more potent than the inter-class connectivities. We calculate the mean connectivity strengths of intra- and inter-classes of the Cora dataset (high connectivity strengths mean more attention weights are concentrated on the edge between the two classes), and visualize the matrix in Fig.~\ref{fig:attention-view}. Because GMNN performs message passing on the original graph, so its connectivity strength is the same as GCN. Meanwhile, the result of feature-based attention model GAT is similar to GCN (GMNN), since it omits the label dependency and lack of supervision. 
However, if we replace the GCN in GMNN with GAT and apply attention to labels, the results become better.
Compared with GAT and GMNN methods, our GDAMN method obtains higher scores for the diagonal part, which implies higher intra-class strength.
Moreover, the information ENT regularizer and hard attention constrained by the KL divergence between our graph prior can provide more supervision and significantly help to reduce the inter-class weights ratio. 

\subsubsection{Retrain with vanilla GCN}
To further verify the effectiveness of the learned aggregation weights, we compare the training loss and test accuracy using a vanilla GCN with different edge weights. The weights are the graph Laplacian computed by the initial adjacent matrix, oracle graph Laplacian by removing all the inter-class edge, and the learned aggregation weights by the proposed GDAMN. As shown in Fig.~\ref{fig:learn-edge-curve}, in the training stage, the GCN with the learned attention weights by our GDAMN achieves a similar convergence rate to the oracle, and significantly faster than the GCN with the original adja gcent matrix. In the test stage, our performance also outperforms the GCN with the original adjacent matrix significantly. In this sense, our method can also be used as a graph learner to refine the edge strength of the aggregation weights, which is compatible with any GNN with more sophisticated structures.

\subsubsection{The effectiveness of stable graph re-weighting}
As the discussion in Section~\ref{section:reweighting}, when re-weighting the graph of E-step, the training of GCN may be unstable due to the variance of sample discrete graph structure from the Gumbel-softmax. To show the superiority of our stable re-weighting in E-step, we tested the node classification performance under different graph sampling numbers $S$, and vary it from 1 to 10 on Cora and Citeseer datasets, and 0 indicates our stable re-weighting. The results are shown in Fig.~\ref{fig:stable}, when increasing the sampling numbers $S$, the performance increases slowly. However, the performance of stable re-weighting is consistently higher than the direct sampling of different $S$, which shows that the stable weights ${\mathbf A}^{\text{stable}}$ in Equation~\eqref{eqn::stable-graph} is more efficient and stable for message passing.

\subsubsection{Analysis of time complexity}
The time complexity of a single
$\operatorname{GNN}^{\mathrm{P}}_{\phi}$ layer is $O(|V|FF' + |E|CC + |Ez|F'F'')$, where $F$, $F'$ and $F''$ are the numbers of input, hidden, and projection feature dimensions, 
$C$ is the class numbers, and $|V|$, $|E|$ and $|E_z|$ are the numbers of nodes, original edges, and the latent structure edges, respectively. The cost for bilinear hard attention is $|E|CC$, and the cost for soft attention and message passing is $|E_z|F'F''$. Since we need two-step attention computation, this complexity is slightly higher than the baseline methods such as Graph Attention Networks (GATs)  \cite{GAT}.
In our experiments, we empirically noted that GDAMN converges in three EM iterations.

\section{conclusion}\label{Sec:6}
In this paper, we have proposed a Graph Decoupling Attention Markov Network to decouple the attention learning procedure on the graph into hard and soft attention. \red{The hard attention learns graph structure based on labels, and the soft attention learns edge weights based on features.
To avoid the influence of insufficient label information in the semi-supervised node classification scenario, we use the EM framework to approximate the label distribution and apply the variational inference to model the uncertainty of structure in the M-step.
Moreover, we impose a novel structure prior and the neighbor mutual information constraint in the learning process to obtain better hard and soft attention}. 
Experiments on five benchmark datasets demonstrate that our model outperforms state-of-the-art methods with more reasonable aggregation weights for both feature and label propagation.


\red{Our work can also be regarded as a general framework that considers the features and labels to learn the graph structure and edge weights. In this work, we use the simple GCN as propagator for the feature and label.}
One of our future work is to combine GDAMN with more complex graph neural networks and apply it to computer vision or natural language processing. \red{\blue{[R2.4]}It is also promising to explore other graph structure priors that are reasonable for the graph-level and edge-level tasks. These priors can extend our work to graph-level and edge-level tasks in the future. 
Further, it is interesting to investigate how to combine effective sampling techniques with our methods to learn subgraph structures to deal with extremely large graphs.
}

\IEEEpeerreviewmaketitle

\bibliographystyle{IEEEtran}
\bibliography{biblio}

\end{document}